\documentclass{article}

\PassOptionsToPackage{compress}{natbib}

\usepackage[preprint]{acl}

\usepackage{times}
\usepackage{latexsym}

\usepackage[T1]{fontenc}

\usepackage[utf8]{inputenc}

\usepackage{microtype}

\usepackage{inconsolata}

\usepackage{graphicx}

\usepackage{enumitem}
\usepackage{graphicx}
\usepackage[most]{tcolorbox}
\newtcolorbox{promptbox}[1][]{
  colback=gray!10,            %
  colframe=gray!50, %
  coltitle=black,             %
  boxrule=0.8pt,
  arc=6pt,
  left=6pt,
  right=6pt,
  top=6pt,
  bottom=6pt,
  fonttitle=\bfseries,
  title=#1,
  fontupper=\ttfamily\raggedright
}
\usepackage{etoc}
\usepackage{tabularx}
\usepackage{array}
\usepackage{booktabs}       %
\usepackage{amsfonts}       %

\title{Beyond Prompt and Parser Engineering: Soft-Prompt Tuning for Automated LLM Evaluation}

\author{Selen Erkan \\
  Aleph Alpha Research \\ Lab1141 \\ \\\And
  Bastian Boll \\
Aleph Alpha Research\\ Lab1141 \\ \\\And
  Kristian Kersting \\
   TU Darmstadt \\ Hessian.AI \\ Lab1141 \\ \\\AND
  Björn Deiseroth \\
  Aleph Alpha Research\\ Lab1141 \\ \\\And
  Letitia Parcalabescu \\
  Aleph Alpha Research \\ Lab1141 \\
  }

\begin{document}

\maketitle

\begin{abstract}

Benchmark scores often misrepresent a large language model's (LLM's)  knowledge, because they rely, e.g., on the model's ability to follow specific formatting requirements. This especially penalizes base models that may %
know the correct answers but lack the ability -- typically introduced in post-training -- to structure them as instructed.
To overcome this, we propose an efficient, automated, and architecture-agnostic method for model evaluation based on soft-prompt tuning. By optimizing only 10 soft-prompt vectors ($\sim$0.0006\% parameters for a 7B model) over a %
short tuning period, we adapt models to specific benchmark formats, closing gaps in format-following and ensuring that underlying knowledge is accurately reflected in benchmark scores. %
We evaluate soft-prompt tuning across 7 models and 11 datasets. Our results show that (a) soft-prompt tuning saturates format-following within 150 steps (${\sim}1200$ samples) while preserving knowledge accuracy, (b) it closely matches strong evaluation-harness baselines on well-studied benchmarks while requiring no manual benchmark-specific prompt or parser design, (c) it improves evaluation signal in settings where baselines are harder to transfer, including long-context datasets and instruction-tuned variants with different output styles, and (d) it produces cleaner outputs that improve agreement between exact match scoring and independent LLM judges. Overall, we propose a lightweight, automated, and flexible protocol based on limited-budget soft-prompt tuning for evaluating LLM knowledge.

\end{abstract}

\section{Introduction}
Large language models (LLMs) have achieved impressive performance across a wide range of tasks, from answering factual questions to multi-step math problems \citep{yang_qwen3_2025, deepseek-ai_deepseek-v32_2025, comanici_gemini_2025}.  To track progress, models are routinely evaluated on a wide range of benchmarks \citep{joshi_triviaqa_2017, cobbe_training_2021, hendrycks_measuring_2021}, %
which require adherence to specific output formats. Yet, format-following capability is often latent in base models, becoming more reliable only after post-training. This raises a key question: do benchmark scores actually measure what a model knows, or are they partly bottlenecked by the model's ability to format answers correctly?

Specifically, benchmarks often mandate rigid output formats, unfairly penalizing models that have the correct knowledge but lack the capacity to adhere to strict formatting requirements \citep{long_llms_2025}. Base models are particularly vulnerable here: lacking instruction-following supervision, they often fail to produce parseable answers, e.g., repeating the question multiple times without answering. %

Existing evaluation harnesses \citep{gu_olmes_2025, gao_framework_2021} try to address this problem by designing benchmark-specific prompts, answer formats, and parsing rules. However, this process requires significant manual effort and validation \citep{liu_pre-train_2023}, and such pipelines may not transfer cleanly when model behavior or benchmark conditions change, or to long-context tasks where few-shot prompting is limited by maximum context length \cite{yen_helmet_2025}. Furthermore, model performance is fragile against prompt variations, meaning even minor wording or format changes can significantly alter benchmark scores \citep{sclar_quantifying_2024, habba_dove_2025}.

A complementary strategy is to use LLM-as-a-judge evaluation to extract the intended answer from noisy model generations. While this can reduce dependence on exact match parsers, it shifts the burden to designing and validating the judge itself. LLM judges are sensitive to their own prompts and can exhibit systematic biases \citep{ye_justice_2025}. \textbf{Thus, existing evaluation pipelines can require substantial model- and benchmark-specific engineering.}

To address this gap, we introduce an \textit{automated}, compute- and memory-efficient evaluation framework based on  soft-prompt tuning (SP) \cite{lester_power_2021}.
Parameter-efficient methods such as soft-prompt tuning are widely used to \textit{improve} task performance. We instead use them to \textit{measure} it. With a small tuning budget  of only $\sim$150 steps on the benchmark's \textit{training} split (corresponding to 1200 data points, and 0.0006\% trainable parameters on a 7B model), each model discovers its own prompt per task, adapting how it expresses answers without learning new task knowledge. To verify this, we propose metrics to decouple \textit{knowledge} from \textit{format-following accuracy} and show that soft-prompt tuning reaches 80–100\% format-following while preserving knowledge accuracy, allowing benchmark scores to more faithfully reflect the model's underlying knowledge.

We validate this approach across a diverse suite of seven models, including Llama-3, Qwen-3, OLMo-3, and Ministral-3, and 11 benchmarks covering factual question answering, reading comprehension, mathematical reasoning, and long-context evaluation. On well-studied benchmarks, e.g., TriviaQA, soft-prompt tuning  closely matches few-shot evaluation harness baselines (0.60 soft-prompt vs 0.59 few-shot score), while providing a more automated alternative avoiding manual prompt and parser design. Furthermore, soft-prompt tuning transfers to settings where manually engineered baselines are brittle or impractical. It strengthens evaluation signal on long-context benchmarks, where few-shot prompting is difficult to use, and improves evaluation of instruction-tuned variants, where chat-template and output-style mismatches might require updated baseline protocols. Moreover, soft-prompt-tuned outputs are clean enough that even strict exact match agrees closely with independent LLM judges, reducing the need for a costly judge-based evaluation setup. They also increase agreement between different judges.
Finally, a small manual inspection shows that soft-prompt tuning can correct parser-missed answers from diverse base models. Our main contributions are:

\begin{enumerate}[label=\roman*), topsep=2pt, noitemsep, leftmargin=*]
    \item We introduce a lightweight soft-prompt tuning-based LLM evaluation protocol that separates knowledge from format-following ability.

    \item To validate our protocol, we propose two complementary metrics, \textit{knowledge} accuracy and \textit{format-following} accuracy, to directly measure whether format adaptation changes benchmark compatibility without injecting new systematic task-specific knowledge. These rely on costly calibrated LLM judges, so we only use them to verify the method, not as part of the evaluation protocol itself.
    
    \item We show that soft-prompt tuning automates benchmark-format adaptation, matching  established evaluation protocols on well-studied benchmarks, yet we do not rely on benchmark-specific prompt design, parser engineering, or extensive LLM-judge calibration.

    \item We show that our protocol outperforms manually engineered baselines where they are brittle: long-context evals, instruction-tuned models, diverse base-model output styles, and judge-based evaluation, where its cleaner outputs improve judge–exact-match and cross-judge agreement.
\end{enumerate}

\section{Related Work}
\label{related_work}
\subsection{Model Evaluation}

\textbf{Limitations of evaluation harnesses.}
Benchmark evaluation requires (1) a prompt that elicits the expected answer format and (2) a procedure that can extract the model's prediction from potentially noisy generations. Evaluation harnesses provide benchmark-specific evaluation pipelines for this, with prompt templates, fixed few-shot examples, answer extraction rules, and scoring metrics \citep{gu_olmes_2025, gao_framework_2021}. These improve reproducibility on well-studied benchmarks but require effort to set up for new models and benchmarks. Since model scores are sensitive to design choices such as prompt wording and format \citep{lu_how_2024, sclar_quantifying_2024, cao_worst_2024} and answer extraction rules, which may favor models whose outputs match the expected format \citep{biderman_lessons_2024, yu_xfinder_2025}, extending these pipelines to new scenarios is non-trivial. Multi-prompt evaluation reduces this sensitivity by aggregating scores over many prompt templates, but still requires a template pool \citep{mizrahi_state_2024, polo_efficient_2024}. Trained answer extractors such as xFinder replace regex parsers, but require large annotated response datasets and are trained on a fixed set of answer types \citep{yu_xfinder_2025}. \citet{bunn_fine-tune_2025} fine-tune models to improve format-following, but this requires expensive full fine-tuning on 32,000 samples and targets multiple-choice tasks, where log-likelihood scoring already works well.

\textbf{Limitations of log-likelihood evaluation}.
For closed-ended tasks such as multiple choice, models are often evaluated by ranking candidate answers by log-likelihood  \citep{hendrycks_measuring_2020, srivastava_beyond_2023}. This avoids format issues but requires a fixed candidate set, so it does not apply to open-ended generation \citep{biderman_lessons_2024}.

\textbf{Limitations of loss based evaluation.} Loss on held-out data is used as a prompt-free proxy for downstream performance \citep{du_understanding_2024, chen_scaling_2025}, but it does not always correlate with generative metrics or real-world usability  \citep{isik_scaling_2024, tay_scale_2021}. %

\subsection{Parameter efficient fine-tuning (PEFT)}

\textbf{Soft-prompt tuning} is a parameter-efficient fine-tuning method \citep{lester_power_2021}:  the model stays frozen, and only a few learnable vectors -- prepended to the context at the embedding level -- are optimized to maximize the likelihood of the correct answer. Unlike hard (text) prompts, soft-prompts are continuous and can be optimized directly via gradient descent. Soft-prompts and related prompt-tuning methods have been studied extensively for task adaptation \citep{li_prefix-tuning_2021, liu_p-tuning_2022}, but rarely as an evaluation mechanism. Prior work has used continuous prompts to probe factual knowledge in BERT-style masked language models (LMs) through single-token fill-in-the-blank queries \citep{qin_learning_2021, zhong_factual_2021}. \citet{qin_learning_2021} do not examine whether such prompts learn from the training data, while \citet{zhong_factual_2021} show that in their setup they partly do. We instead apply soft-prompts to generative LLMs on open-ended benchmarks: each model discovers its own prompt on the benchmark's train split, adapting its output format at a fraction of the cost of full post-training. We constrain tuning to limit task learning and explicitly measure knowledge separately from format-following, so that benchmark scores better reflect what the model knows.

\textbf{Other PEFT methods} \citep{liu_few-shot_2022, dettmers_qlora_2023} include adapters \citep{houlsby_parameter-efficient_2019}, which insert small trainable layers  between the frozen layers, but add inference latency. Low-Rank Adaptation (LoRA) \citep{hu_lora_2021} learns low-rank updates to the frozen weights. \citet{zhang_train-before-test_2025} apply LoRA fine-tuning before evaluation to stabilize model rankings \textit{across datasets}, but neither compares  base and post-trained models nor disentangles format from knowledge.

We use soft-prompt tuning due to its simplicity and computational efficiency. We repeat our core experiments with LoRA (Appendix~\ref{lora_experiments_appendix}) and find similar trends, but LoRA trains far more parameters (0.27\% vs.\ 0.0006\% for a 7B model). %

\section{Soft-Prompt Tuning for Automated LLM Evaluation}
\label{sec:method}
\label{findings}
In the following, we introduce soft-prompt tuning and discuss how to use it for fair benchmarking.

\subsection{Background: Soft-Prompt Tuning}
\label{sec:soft_prompt_background}
Soft-prompt tuning forgoes discrete  natural-language instructions in favor of continuous embedding vectors \citep{lester_power_2021}. Given a frozen LLM with parameters $\theta$ and embedding dimension $d$, we define a soft-prompt as a sequence of $L$ learnable vectors $\mathbf{S} = \{s_1, s_2, \dots, s_L\}$, where each $s_i \in \mathbb{R}^d$.

For an input sequence $X$ with token embeddings $E(X)$, the input to the first transformer layer is the concatenation $Z = [\mathbf{S}; E(X)]$. The model then generates the target sequence $Y = \{y_1, y_2, \dots, y_m\}$ autoregressively. The probability of the sequence is
\begin{equation}
    P_\theta(Y \mid \mathbf{S}, E(X)) = \prod\nolimits_{t=1}^{m} P_\theta(y_t \mid \mathbf{S}, E(X), y_{<t}).
\end{equation}
Model weights $\theta$ remain frozen during tuning, preserving the model's latent knowledge. We optimize only the soft-prompt $\mathbf{S}$ by minimizing the negative log-likelihood over the benchmark's \textit{training} split $\mathcal{D}_{\text{train}}$
\begin{subequations}
\begin{align}
\hat{\mathbf{S}}
&= \arg\min_{\mathbf{S}}
\sum_{(X,Y)\in\mathcal{D}_{\text{train}}} \mathcal{L}(\mathbf{S}; X,Y), \\
\mathcal{L}(\mathbf{S}; X,Y)
&= -\sum_{t=1}^{m}
\log P_\theta(y_t \mid \mathbf{S}, E(X), y_{<t}).
\end{align}
\end{subequations}
At evaluation time, $\hat{\mathbf{S}}$ is prepended to each input of the held-out test split, which is never seen during tuning.

\subsection{Soft-Prompts as a Limited-Budget Format Adapter}
\label{method_desc}

We argue that soft-prompt tuning is well-suited to disentangle what a model knows from how well it follows benchmark formats, for two reasons:

\textit{(i) Limited-budget tuning.} We use $L = 10$ soft-prompt vectors ($\sim 0.0006\%$ of a $7$B model's parameters), $T = 150$ optimization steps, and $1200$ total training samples per benchmark. We hypothesize that this budget suffices to adapt output format but not to encode substantial benchmark-specific knowledge, and test this empirically in the experiments section (Sec.~\ref{experiments}).

\textit{(ii) Self-discovered prompts.} Unlike hard prompts, soft-prompts are optimized directly in embedding space against the benchmark's train split. Each model thus  discovers its own prompt for each benchmark, bypassing the effort and limitations of prompt engineering. %

Both properties (i) and (ii) motivate our main hypotheses H1 and H2 that 
we will evaluate in Sec.~\ref{sec:experiments}:

\textbf{H1: Format without knowledge.} Within the limited tuning budget, soft-prompt tuning saturates format-following while leaving knowledge accuracy largely unchanged, so that benchmark scores better reflect what the model knows.

\textbf{H2: Automated and flexible benchmark adaptation.} Because each model discovers its own prompt, soft-prompt tuning adapts models to benchmark formats without manual prompt search, task-specific parsers, or LLM-judge calibration. It should therefore match strong hand-engineered baselines on well-studied benchmarks, and remain useful where such baselines are brittle (diverse base-model output styles, instruction-tuned models, long-context benchmarks where few-shot prompting is costly), and and yield answers clean enough for strict exact match to align with complex LLM judges.

\subsection{Metrics for Disentangling Knowledge from Format-Following (H1)}
\label{sec:metrics}

To test \textbf{H1} during soft-prompt tuning, we introduce two complementary metrics: \textit{knowledge accuracy} and \textit{format-following accuracy}. We compute both metrics differently for open-ended and closed-ended (multiple-choice) datasets, as described below. Closed-ended datasets serve as a controlled setting for validating the patterns observed on open-ended datasets. For the remainder of the paper, we primarily focus on open-ended datasets, where the limitations described above are most pronounced.

Because these metrics rely on calibrated LLM judges that are costly to engineer for each new model and benchmark, we use them only at research time to validate the method. The soft-prompt evaluation protocol itself does not require them; avoiding this manual effort is exactly its purpose.

\textbf{Closed-ended datasets.} \textit{Knowledge accuracy} uses standard log-likelihood scoring: the candidate answer with the highest log-likelihood is the prediction. Since no free text is generated, this isolates knowledge from format-following. \textit{Format-following accuracy} measures whether the model's free-text generation exactly matches one of the answer options (e.g., ``A.\ Berlin''), regardless of correctness. \ref{appendix_metric_examples} shows examples of both.

\textbf{Open-ended datasets.} We use Qwen-3-32B \citep{yang_qwen3_2025} as an LLM judge for both metrics. For \textit{knowledge accuracy}, it checks whether the response contains the correct answer, regardless of formatting; for \textit{format-following accuracy}, whether the response matches the ground truth's format, regardless of correctness. Both scores are binary. We refined the judge prompts on 30 human-labeled examples and validated them on 100 further samples, reaching 84\% agreement with human annotations (prompts in Appendix~\ref{appendix_llm_judge}).

Unlike log-likelihood scoring, LLM judges are sensitive to prompt design and prone to systematic biases, so open-ended knowledge accuracy is not a fully disentangled measure. We therefore use closed-ended datasets as the controlled setting for validating H1, and treat open-ended results as supporting evidence.

\subsection{Soft-Prompts as an Automated Evaluation Protocol (H2)}
\label{sec:h2_protocol}
\textbf{H1} establishes that, under a limited budget, soft-prompt tuning changes how a model expresses its answers rather than what it knows. \textbf{H2} asks whether this makes it a practical evaluation protocol.

Tuning takes only 1--2 minutes per model--benchmark pair on a single GPU, and, unlike hand-engineered prompts, parsers, or judges, needs no redesign for new models or benchmarks. We test this in five settings: against strong baselines on well-studied benchmarks, across base and instruction-tuned models, long-context tasks, LLM-judge agreement, and diverse base-model output styles.

\section{Experimental Evaluation of Soft-Prompt Tuning}
\label{sec:experiments}
\label{experiments}

We conduct experiments with seven models across eleven benchmark datasets, covering both closed-ended (multiple-choice) and open-ended (question-answering short and long-context) tasks as outlined below.

\subsection{Experimental Setup} \label{sec:exps-setup}

\textbf{Models.} In the experiments we used seven models: Llama-3.1-8B, Llama-3.2-1B \citep{grattafiori_llama_2024}, Ministral-3-14B, Ministral-3-3B \citep{liu_ministral_2026}, Olmo-3-7B \citep{olmo_olmo_2026}, Qwen-3-8B, and Qwen-3-4B \citep{yang_qwen3_2025} base models and their post-trained counterparts. To evaluate open-ended datasets, we used Qwen-3-32B as LLM-as-a-judge and used Olmo-2-32B \citep{olmo_2_2025} to test how well different LLM-judges align with each other.

\textbf{Datasets.} We used five open-ended, three long-context and three closed-ended English datasets. As open-ended datasets we used TriviaQA, GSM8K, SQuAD, DROP and MATH. TriviaQA measures factual question-answering capabilities and requires models to provide short answers without explanations \citep{joshi_triviaqa_2017}. GSM8K measures mathematical reasoning on grade-school problems, requiring the model to produce the correct final numerical answer along with brief reasoning \citep{cobbe_training_2021}. SQuAD evaluates reading comprehension by requiring the model to extract the precise answer to a given question directly from a provided context paragraph \citep{rajpurkar_squad_2016}. DROP evaluates discrete reasoning over paragraphs, requiring models to perform operations such as counting, comparison, and arithmetic over information in the passage \citep{dua_drop_2019}. Finally, MATH evaluates mathematical problem solving on competition-style questions that require multi-step reasoning and a final answer in a structured mathematical format \citep{hendrycks_measuring_2021}. See App.~\ref{appendix_benchmark_prompts} for benchmark examples and prompts. 

For the long-context datasets we used QASPER, QuALITY, ContractNLI benchmarks. QASPER asks questions about long scientific papers and requires the model to find the answer in the paper \citep{dasigi_dataset_2021}. QuALITY tests whether the model can answer questions about long passages or stories \citep{pang_quality_2022}. ContractNLI uses legal contracts and asks the model to decide whether a statement is supported by, contradicted by, or not found in the contract \citep{koreeda_contractnli_2021}. All three are evaluated by free-form generation: QASPER has free-form answers, whereas QuALITY and ContractNLI require generating one of a fixed set of answers.

The closed-ended datasets are MMLU, Belebele, and ChemBench, all of which are multiple-choice benchmarks. MMLU covers 57 tasks including elementary mathematics, US history, computer science, law, and more \citep{hendrycks_measuring_2020}. Belebele evaluates reading comprehension \citep{bandarkar_belebele_2024}, and ChemBench focuses on undergraduate- and graduate-level chemistry questions \citep{mirza_framework_2025}. In all closed-ended datasets, the answer options are explicitly provided to the model. The model must output the correct answer in the required format, which usually includes the answer option (when available) and its full text. For MMLU, we present answer options as full text without letter identifiers, so the expected output is the correct option text itself.

We use the original dataset splits and construct an unseen 800-example development set from each training split. For datasets without an official split we created train/dev/test splits. DROP and MATH test evaluations are limited to 1000 examples. We tune on train, select the optimum tuning steps on dev, and report results on test. 

\textbf{Metrics.} We evaluate models using complementary metrics defined in Sec.~\ref{sec:method}: (i) \textit{format-following accuracy}, (ii) \textit{knowledge accuracy} and (iii) \textit{exact match accuracy} on an autoregressive text generation task used uniformly across open- and closed-ended datasets with standard benchamrk normalization. These metrics test H1 (Format without Knowledge) and H2 (Automated and Flexible Benchmark Adaptation) separately, and track changes in model knowledge and format-following across soft-prompt tuning and the baselines. %

\textbf{Baselines.} Evaluation harnesses (Sec.~\ref{related_work}) provide strong standardized alternatives to our method, but they typically rely on benchmark-specific prompt and parser design. We compare against the zero- and few-shot variants of the OLMES evaluation protocol \citep{gu_olmes_2025} to test whether soft-prompt tuning can provide a more automated and flexible evaluation protocol.

\textbf{Tuning details.} For all experiments (all datasets and models), we used $L=10$ soft-prompt vectors (App.~\ref{softprompt_vector_ablation}, Fig.~\ref{fig:sp_vector_ablation} for vector-size ablation), learning rate 0.001,
a batch size of 8 and the AdamW optimizer \citep{loshchilov_decoupled_2018}.
We initialize soft-prompt embeddings $\mathbf{S} \in \mathbb{R}^{L \times d}$ following a normal distribution $\mathcal{N}(0, \sigma^2)$, $\sigma = 0.02$ and tune them using one GPU on the benchmark's training split.

\subsection{Results}
We present the empirical results addressing the two hypotheses from Sec.~\ref{method_desc}.

\subsubsection{Soft-prompt tuning maximizes format-following on both base and post-trained models (validating H1)} 
\label{soft_prompt_format_following}
\begin{figure}[t]
\centering
\raggedright
\makebox[0.9\linewidth][l]{\textnormal{(a)}}\\[-2pt]
\includegraphics[width=0.8\linewidth]{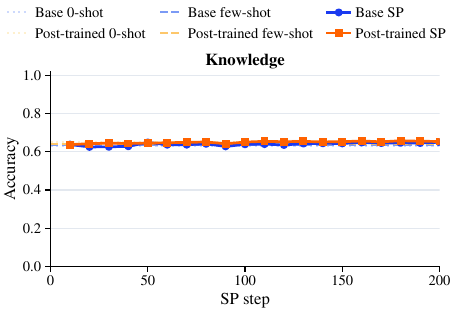}
\vspace{4pt}
\raggedright
\makebox[0.9\linewidth][l]{\textnormal{(b)}}\\[-2pt]
\includegraphics[width=0.8\linewidth,trim=0 0 0 20, clip]{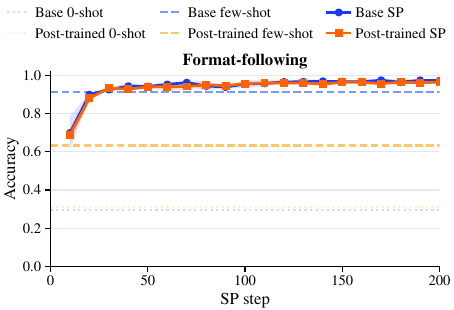}
\caption{\textbf{Soft-prompts maximize format-following and do not show additional knowledge gains} --  \textit{closed-ended datasets.} (a) \textit{Knowledge} and  (b) \textit{format-following} metrics across 200 tuning steps. Results are aggregated over 3 closed-ended datasets, 7 models and 3 seeds.}%
\label{fig:closed_ended_metrics_fig}
\end{figure}

\begin{figure}[t]
\centering
\raggedright
\makebox[0.9\linewidth][l]{\textnormal{(a)}}\\[-2pt]
\includegraphics[width=0.8\linewidth]{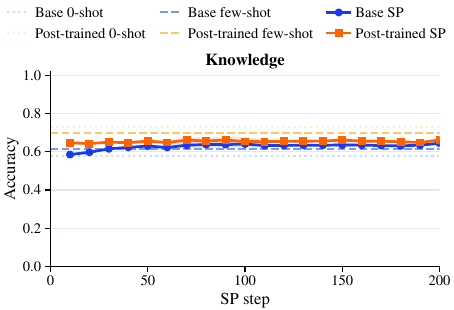}
\vspace{4pt}
\raggedright
\makebox[0.9\linewidth][l]{\textnormal{(b)}}\\[-2pt]
\includegraphics[width=0.8\linewidth,trim=0 0 0 20, clip]{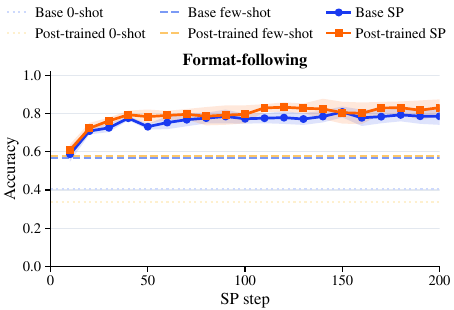}
\caption{\textbf{Soft-prompts maximize format-following and do not show knowledge metric gains} --  \textit{open-ended datasets.} (a) \textit{Knowledge} and  (b) \textit{format-following} metrics across 200 tuning steps. All results are aggregated over 5 open-ended datasets, 7 models and 3 seeds}%
\label{fig:open_ended_metrics_fig}
\end{figure}

Figs.~\ref{fig:closed_ended_metrics_fig} \& \ref{fig:open_ended_metrics_fig} show knowledge and format-following accuracy on the dev split for \textit{closed}- and \textit{open-ended} datasets. Curves represent aggregates over datasets and models, shaded regions indicate variance across 3 random seeds. Per dataset and model results are in \ref{appendix_closed_ended_knowledge_format} \& \ref{appendix_open_ended_knowledge_format}.

On closed-ended datasets (Fig.~\ref{fig:closed_ended_metrics_fig}b), format-following accuracy approaches 100\% by around step 40. This fast convergence is expected: the required format is simple, typically copying one of the provided options.

On open-ended datasets (Fig.~\ref{fig:open_ended_metrics_fig}b), judge-measured format-following accuracy reaches 80--100\% around step 150. Majority of the per benchmark and model curves (\ref{appendix_closed_ended_knowledge_format} \& \ref{appendix_open_ended_knowledge_format}) show similar saturation by this point, so we use step 150 for all subsequent experiments.

Knowledge accuracy, in contrast, stays largely flat over all 200 steps on both closed-ended (Fig.~\ref{fig:closed_ended_metrics_fig}a) and open-ended datasets (Fig.~\ref{fig:open_ended_metrics_fig}a). Together with the format-following curves, this supports H1: within our budget, soft-prompt tuning saturates format-following without measurably adding task knowledge.

Manual inspection also reveals cases where the judge conflates knowledge with format-following (App.~\ref{llm_judge_failure_examples}), which reinforces why we rely on closed-ended datasets as the controlled test of H1.

\textbf{Post-trained models also benefit.} 
Despite instruc\-tion-tuning, post-trained models (orange in Figs.~\ref{fig:closed_ended_metrics_fig}b and~\ref{fig:open_ended_metrics_fig}b) follow benchmark formats poorly under zero- and few-shot prompting. Soft-prompt tuning raises them to the same format-following level as base models.

\subsubsection{Soft-prompt tuning as an automated evaluation protocol (validating H2)}
Next, we evaluate whether soft-prompt tuning can serve as an automated evaluation protocol. We test H2 in five settings: well-studied benchmarks, instruction-tuned models, long-context tasks, LLM-judge agreement, and diverse base-model output styles.

\textbf{Soft-prompt tuning automates evaluation on established benchmarks.}
Well-studied benchmarks such as TriviaQA \citep{joshi_triviaqa_2017} have mature evaluation pipelines, with carefully designed prompts and parsers that are often able to extract model knowledge. We test whether our automated soft-prompt evaluation can recover similar signals without benchmark-specific manual engineering. We evaluate 7 base models on TriviaQA, GSM8K, SQuAD, MATH, and DROP with zero-shot, few-shot baselines, and soft-prompt evaluation.

\begin{figure}[t]
\centering
\raggedright
\textnormal{(a)}\\[-2pt]
\includegraphics[width=0.9\linewidth]{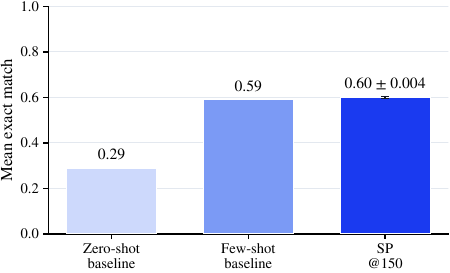}
\vspace{4pt}
\raggedright
\makebox[0.9\linewidth][l]{\textnormal{(b)}}\\[-2pt]
\includegraphics[width=0.9\linewidth]{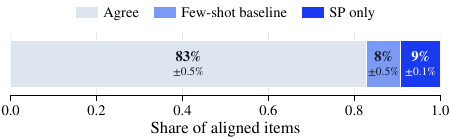}
\caption{\textbf{Automated soft-prompt evaluation matches well-studied benchmark scores} -- \textit{open-ended datasets}. (a) Aggregated exact match metric. (b) Aggregated item-level alignment between soft-prompt tuning and the few-shot baseline. Results are aggregated over 5 open-ended benchmarks, 7 models, and 3 seeds. Soft-prompt tuning results report the mean across 3 seeds, with error bars showing variation across seeds.}
\label{fig:base_sp_olmes_align}
\end{figure}

Soft-prompt tuning closely matches the few-shot baseline in exact match (0.60 vs.\ 0.59), while zero-shot lags far behind (0.29; Fig.~\ref{fig:base_sp_olmes_align}a). The two methods also agree on $\sim$83\% of individual items (Fig.~\ref{fig:base_sp_olmes_align}b). This alignment at both aggregate and item level, achieved without benchmark-specific prompts or parsers, shows that soft-prompt tuning can serve as an automated alternative to established protocols, and is consistent with it not measurably adding task knowledge.

The remaining disagreements split evenly (8\% solved only by few-shot, 9\% only by soft-prompt tuning), so neither method dominates. Manual inspection (App.~\ref{app:few_shot_sp_reasoning_examples}) shows opposite failure modes: few-shot outputs tend to be too long, adding extra context, spurious ``Question:''/``Answer:'' turns, or symbols such as ``\%'' that exact match rejects, while soft-prompt outputs tend to be too short, omitting part of the answer or returning one element of a multi-part answer. On GSM8K and MATH, both methods show similar reasoning failures: reporting an intermediate result, stating a final answer that differs from the one computed, or making arithmetic errors.

\begin{figure}[t]
\centering
\raggedright
\textnormal{(a)}\\[-2pt]
\includegraphics[width=0.85\linewidth]{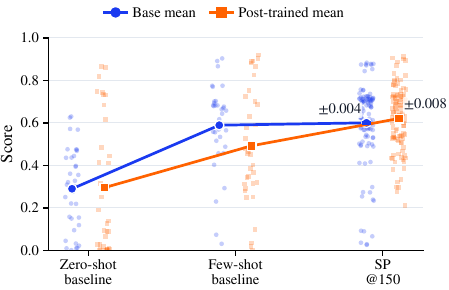}
\vspace{4pt}
\raggedright
\makebox[0.9\linewidth][l]{\textnormal{(b)}}\\[-2pt]
\includegraphics[width=0.85\linewidth]{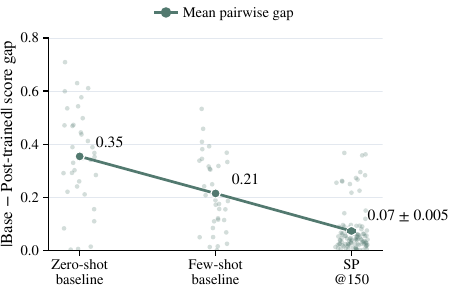}
\caption{\textbf{Soft-prompt tuning reduces the gap between base and instruct variants} -- \textit{open-ended datasets.} (a) Aggregated benchmark scores for base and instruction-tuned models under zero-shot, few-shot, and soft-prompt evaluation. (b) Average pairwise performance gap between each base model and its instruction-tuned counterpart across evaluation methods. Soft-prompt tuning results report the mean across 3 seeds, with $\pm$ values showing the deviation across seeds.}
\label{fig:instruct_base_gap}
\end{figure}

\textbf{Soft-prompt tuning closes the gap between base and instruct model variants and provides better signal on both model variants.}
Comparing base and instruction-tuned variants on the same benchmarks can reveal how post-training changes a model, but only if the evaluation protocol works for both. While existing zero- and few-shot baselines already make some choices that are specific to instruction-tuned models, such as different stop tokens or chat style formatting, they remain largely engineered for base-model behavior and may not transfer cleanly to instruction-tuned variants.
To test this, we evaluate instruction-tuned variants of 7 models on the 5 benchmarks: TriviaQA, SQuAD, GSM8K, MATH, and DROP. We compare soft-prompt tuning against the zero- and few-shot baselines for these benchmarks. 
For instruction-tuned models, soft-prompt tuning clearly outperforms both baselines (0.63 vs.\ 0.48 few-shot), while for base models it stays close to few-shot (0.60 vs.\ 0.59; Fig.~\ref{fig:instruct_base_gap}a). Soft-prompt tuning thus preserves the base-model signal while adapting better to instruction-tuned models, where evaluation protocols engineered for base models transfer poorly.

As a result, the mean gap between each base and instruction-tuned model shrinks from 0.35 (zero-shot) to 0.21 (few-shot) to 0.07 under soft-prompt tuning (Fig.~\ref{fig:instruct_base_gap}b), consistent with findings that suitably prompted base models can match instruction-tuned ones \citep{munjal_instruction-tuned_2026}. Importantly, the gap shrinks while both scores are high, so it reflects knowledge surfaced in both models rather than scores collapsing together. Per model and benchmark results are in App.~\ref{appendix_benchmark_acc}.

\textbf{Soft-prompt tuning supports evaluation when few-shot prompting is impractical.}
In long-context benchmarks, few-shot examples are difficult to use \citep{cattan_doubledipper_2025}. Each input can already occupy a large portion of the context window, so adding multiple long demonstrations may exceed the available context length or leave too little space for the actual test instance. We evaluate soft-prompt tuning on three long-context benchmarks, QASPER, QuALITY, and ContractNLI, to test whether it can provide a useful evaluation signal. As baselines, we use the zero-shot prompts and parsers implemented in the OLMES evaluation suite \citep{gu_olmes_2025} following the standard generative task protocol \citep{shaham_scrolls_2022, shaham_zeroscrolls_2023}. These tasks have no default few-shot demonstrations but instead include task instructions that specify the expected output (e.g., listing the valid labels, or asking for a short phrase or ``unanswerable''). For soft-prompt tuning, we remove these instructions and keep only the input fields (App.~\ref{long_context_prompts}), so each model must learn the expected answer space and style from the training split alone.%

\begin{figure}[t]
\centering
\raggedright
\textnormal{(a)}\\[-2pt]
\includegraphics[width=0.85\linewidth]{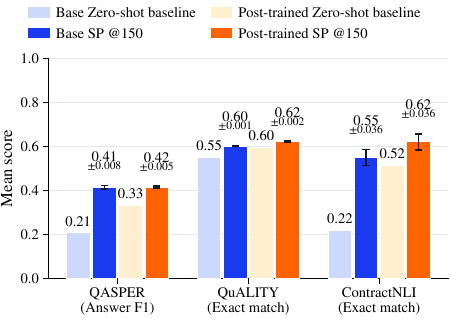}
\vspace{4pt}
\raggedright
\makebox[0.9\linewidth][l]{\textnormal{(b)}}\\[-2pt]
\includegraphics[width=0.85\linewidth,trim=0 0 0 20, clip]{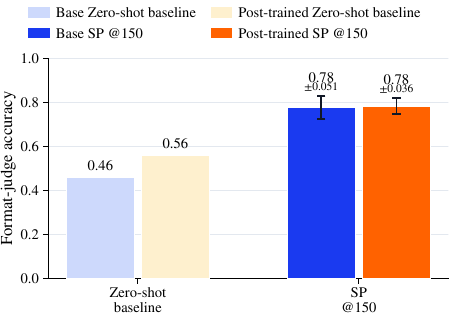}
\caption{\textbf{Soft-prompt tuning provides better signal on long-context datasets}. (a) Aggregated benchmark accuracy (exact match / F1 metric). (b) Aggregated format-following score for zero-shot baseline and soft-prompt tuning at step 150. Results are aggregated over 3 long-context benchmarks, 7 models and 3 seeds. Soft-prompt tuning results report the mean across 3 seeds, with error bars showing variation across seeds.}
\label{fig:longcontext}
\end{figure}

Figure~\ref{fig:longcontext}a shows that soft-prompt tuning yields higher benchmark scores compared to the baseline for both base and instruction-tuned models across all three benchmarks. These higher scores coincide with a large rise in format-following accuracy (Fig.~\ref{fig:longcontext}b) indicating that soft-prompt tuning provides better benchmark signal on the long-context datasets. This mirrors the behavior observed on short-context benchmarks, soft-prompt tuning improves the evaluation signal by adapting model outputs to the benchmark-specific answer format.

For QASPER, we report F1 for consistency with the baseline protocol. Under exact match metric, the zero-shot baseline scores near 0\%, whereas soft-prompt tuning reaches 21\%, showing that it also improves the signal on stricter metrics. Format-following accuracy on QASPER remains lower than on the other long-context benchmarks ($\sim$53\%, App. Fig.~\ref{fig:sp_long_context_format_appendix}), likely because its answers vary in length and type, making the expected format less clear for the judge unlike QuALITY and ContractNLI. We report representative failures in App.~\ref{llm_judge_failure_examples}. Larger soft-prompt budget, longer tuning or multi-judge evaluation can be promising directions to narrow this gap. Nevertheless, soft-prompt tuning still provides a stronger format-following and benchmark signal than the zero-shot baseline, even under our default lightweight setup.

\textbf{Soft-prompt tuning improves agreement between exact match and LLM-judge evaluation.}
To study LLM-judge noise, we evaluate few-shot and soft-prompt outputs with two judges, Qwen-3-32B and OLMo-2-32B. We measure two forms of agreement. First, we compute agreement between each judge and exact match scoring, which indicates whether the generated outputs are clean enough that a strict exact match and a judge recover the same answer. Second, we compute agreement between the two judges, which measures how robust the evaluation is to different judge models. Higher agreement on both measures suggests that the model outputs are less ambiguous and easier to evaluate. 

\begin{figure}[t]
\centering
\raggedright
\textnormal{(a)}\\[-2pt]
\includegraphics[width=0.95\linewidth]{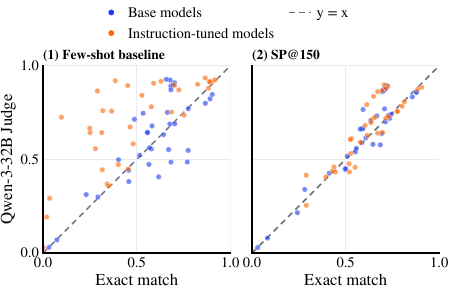}
\vspace{4pt}
\raggedright
\makebox[0.9\linewidth][l]{\textnormal{(b)}}\\[-2pt]
\includegraphics[width=0.95\linewidth,trim=0 0 0 20, clip]{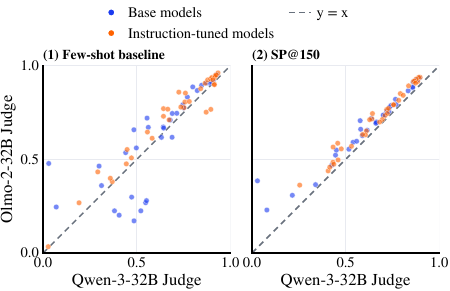}
\caption{\textbf{Soft-prompt tuning better aligns exact match and LLM-judge} -- \textit{open-ended datasets}. (a) Agreement between Qwen-3-32B judge and exact match scores. (b) Agreement between Qwen-3-32B and OLMo-2-32B judges. Both evaluated on (1) few-shot baseline answers and (2) soft-prompt tuned model answers. Each point represents one model--benchmark score; soft-prompt points are averaged over 3 seeds.}
\label{fig:llm_judge_alignment}
\end{figure}

Figure~\ref{fig:llm_judge_alignment} a) shows that Qwen-3-32B and exact match scores align best under soft-prompt tuning for both base and instruction-tuned models. Figure~\ref{fig:llm_judge_alignment} b) shows the same pattern for inter-judge agreement between Qwen-3-32B and OLMo-2-32B. We report the OLMo--exact-match alignment in Appendix~\ref{appendix_olmo_alignment} and observe the same trend as for Qwen--exact-match alignment. These results support our claim that soft-prompt tuning encourages models to produce well-formatted, less ambiguous answers, making exact match accuracy reliable on open-ended benchmarks and reducing the need for LLM judges or manually engineered parsers.

\textbf{Soft-prompt tuning reduces parser failures across diverse base models.}
Using DROP, TriviaQA, and SQuAD, where soft-prompt tuning provides most gains over the baselines, we manually identify baseline failure cases and measure how often soft-prompt tuning corrects them. Base models can express the same correct answer in different ways, especially because they have not been tuned to follow instructions. Fixed prompts and parsers may therefore work well for some models but fail to recognize valid answers from others. In contrast, soft-prompt tuning adapts separately to each model-benchmark pair, allowing each model to express its answers in the expected benchmark format.

\begin{figure}[t]
\centering
\raggedright
\textnormal{(a)}\\[-2pt]
\includegraphics[width=0.85\linewidth]{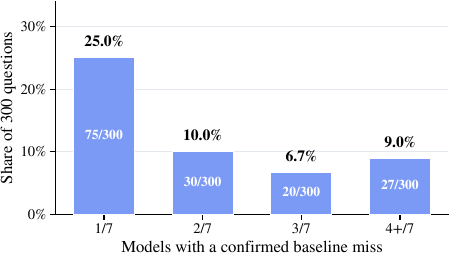}
\vspace{4pt}
\raggedright
\makebox[0.9\linewidth][l]{\textnormal{(b)}}\\[-2pt]
\includegraphics[width=0.85\linewidth]{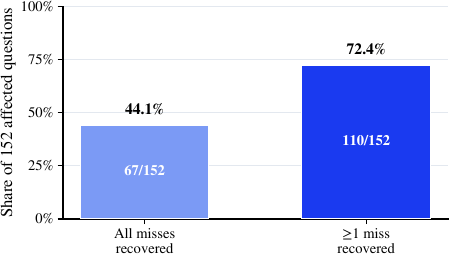}
\caption{\textbf{Soft-prompt tuning can reduce baseline mis-scoring errors} -- \textit{DROP, TriviaQA, and SQuAD datasets.} (a) Percentage of questions for which the baseline evaluation mis-scores correct answers from exactly 1, 2, 3, or 4+ models. (b) Among questions with at least one baseline mis-scoring error, percentage for which soft-prompt tuning corrects all mis-scored answers or at least one mis-scored answer.}
\label{fig:recover_missed_olmes_evals}
\end{figure}

We sample 300 questions (100 per benchmark) for which the 7 base models receive diverse exact match scores under the few-shot baseline evaluation. For each question, we manually inspect 7 model outputs and determine whether score differences reflect genuine knowledge differences or parser failures. In $\sim$50\% of the questions, at least one correct answer is marked incorrect by the baseline parser (Fig.~\ref{fig:recover_missed_olmes_evals}a). Among these 152 questions, soft-prompt tuning corrects the answers of \textit{all} affected models in 44\% of cases (Fig.~\ref{fig:recover_missed_olmes_evals}b). This suggests that soft-prompt tuning recovers part of the signal missed by manually engineered evaluation pipelines. %

\textbf{Additional experiment: global soft-prompt.} We also test whether a \textit{global} soft-prompt per model can replace model-dataset specific prompts. \ref{global_sp_appendix} shows a single 30-token \textit{global} soft-prompt per model, trained on a mixture of 5 open-ended datasets, stays within $\sim$1.5 exact match points of dataset-model specific prompts (tested on SQUAD, TriviaQA and DROP datasets and 3 models). Since dataset-model specific tuning takes only 1--2 minutes, the compute saved is minor, but the results show a promising future direction.

\section{Conclusions}

\label{sec:conclusion}

We introduce soft-prompt tuning as a lightweight, automated protocol for evaluating LLMs. With  limited-budget tuning of 10 soft-prompt vectors, $\sim$0.0006\% of a 7B model's trainable parameters, and 150 tuning steps, soft-prompts adapt model outputs to benchmark formats, reaching 80--100\% format-following accuracy while preserving knowledge accuracy. This shows that soft-prompt tuning improves benchmark compatibility without adding measurable knowledge.

We further compared our protocol to evaluation-harness-style baselines, which rely on manually engineered prompts and parsers for each benchmark. Across 7 models and 11 datasets, soft-prompt tuning closely aligns with existing baselines on well-studied datasets such as TriviaQA, while providing a more automated setup. We also show that soft-prompt tuning extends to scenarios where existing baselines are harder to transfer. It provides stronger signal on long-context benchmarks where few-shot prompting is impractical, and improves evaluation of instruction-tuned variants whose output styles differ from base models. Soft-prompt tuning also encourages models to commit to a single answer and present it in a cleaner benchmark-compatible format, leading to higher judge--exact-match and cross-judge agreement than other baselines. Finally, our manual inspection shows that soft-prompt tuning can correct parser-missed answers from diverse base models.

\section*{Limitations}

We propose a lightweight and automated protocol for LLM evaluation based on constrained soft-prompt tuning. Instead of hand-crafting prompts and parsers for each benchmark, each model learns its own prompt from the benchmark's train split in only 1--2 minutes, using 10 soft-prompt vectors and 150 tuning steps. This budget saturates format-following while leaving knowledge accuracy largely unchanged, so that benchmark scores better reflect what a model knows. Without any manual adaptation, the protocol matches few-shot baseline on well-studied benchmarks and transfers to settings where they are harder to apply, including long-context benchmarks, instruction-tuned models, and judge-based evaluation. Nevertheless, our study has some limitations.

\textbf{Training split requirement.} Our protocol requires a benchmark training split, which not every benchmark provides. When no predefined split exists, constructing one means reporting results on a subset that differs from the official evaluation set. Future work could test whether soft-prompts tuned on a related benchmark with the same answer format transfer, removing the need for an in-domain split.

\textbf{Reliance on LLM judges.} For open-ended datasets, our knowledge and format-following metrics rely on an LLM judge, which can be prompt-sensitive and biased despite our validation. To reduce such errors, we iteratively refined the judge prompts on 30 human-labeled examples and validated them on 100 further samples, reaching 84\% agreement with human annotations. These metrics are used only to validate the method and are not part of the evaluation protocol itself. Their reliability could be further improved by aggregating multiple judges and expanding the human-annotated validation set, although building a reliable judge remains non-trivial.

\textbf{Scope of models and languages.} Our experiments cover only English benchmarks and models up to 14B parameters. Since soft-prompt tuning trains only a few vectors, scaling to larger models would be computationally feasible, and future work can extend the protocol to larger models, further model families, and multilingual benchmarks.

\textbf{Tuning budget in harder settings.} On long-context benchmarks and math datasets with long reasoning traces, format-following remains less saturated, although soft-prompt tuning still outperforms the baselines. We did not systematically study larger soft-prompt budgets or longer tuning schedules for these settings. Future work can explore whether additional tuning capacity closes this gap, while verifying with our knowledge and format-following metrics that the larger budget does not introduce task knowledge.

\bibliography{soft_prompt_references}

\newpage
\appendix
\section{Appendix}
\addcontentsline{toc}{section}{Appendix}

\etocsettocstyle{\subsection*{Contents of Appendix}}{}
\etocsetnexttocdepth{3} %

\begingroup
\parskip=0pt %
\localtableofcontents
\endgroup

\vspace{1em}
\hrule
\vspace{1em}

\subsection{Closed-Ended Datasets Knowledge and Format-Following Accuracy Plots}
\label{appendix_closed_ended_knowledge_format}

\begin{figure*}[t]
\begin{center}
\includegraphics[width=0.7\linewidth]{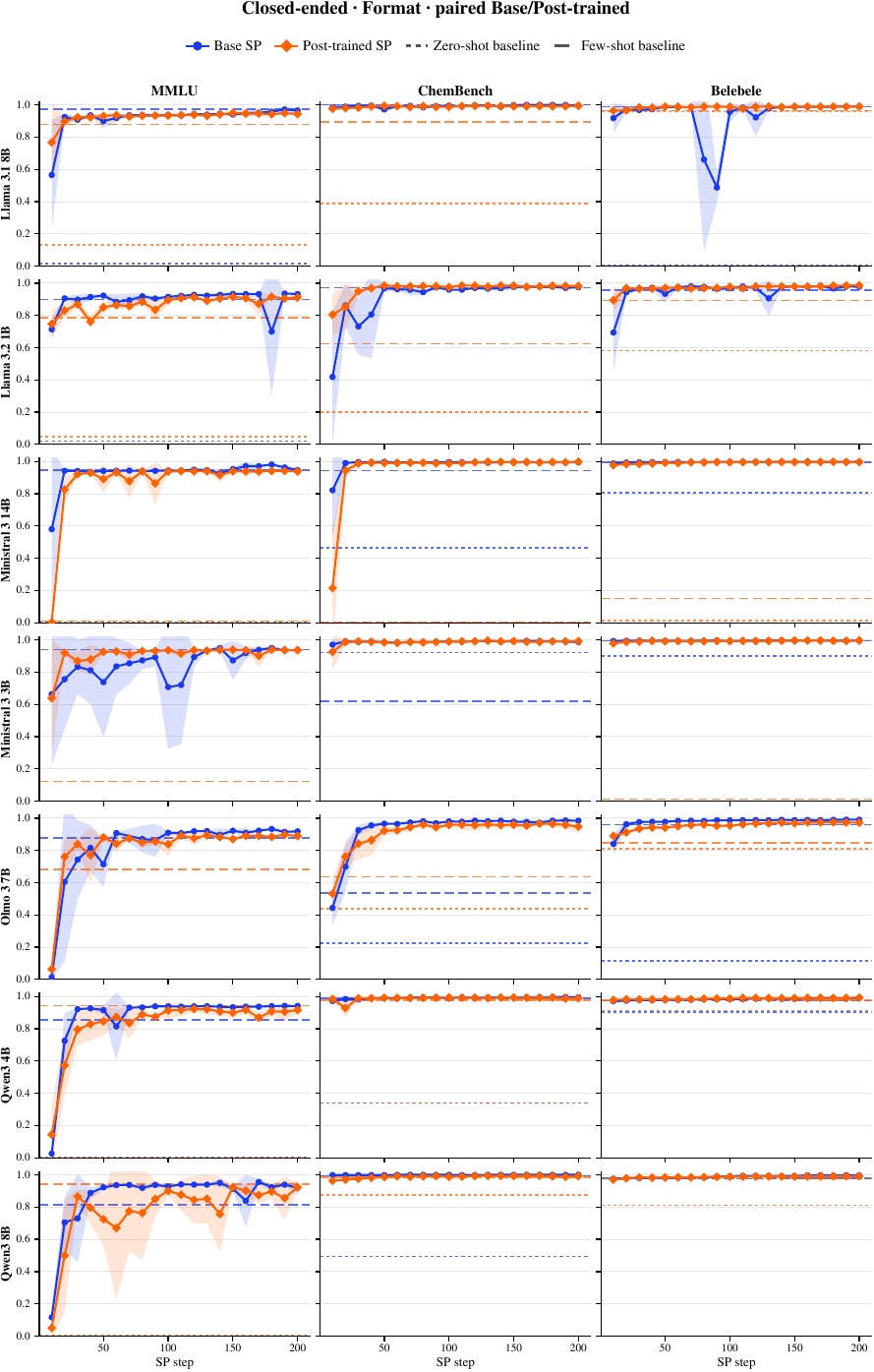}
\end{center}
\caption{\textbf{Soft prompt tuning maximizes format-following -- closed-ended datasets.} Soft-prompt tuning shows 90-100\% format-following accuracy in all plots matching or outperforming the zero- and few-shot baselines. Only in the MMLU plots we see a slight variation. We believe this is because the format-following setup in MMLU dataset is relatively more challenging compared to the other closed-ended datasets. For other datasets model is given option with letters (A. B. etc.) and the full text answers, whereas for MMLU we provide only the full text answer options without any letter options. We believe this way we capture more variety in our experiments. Shaded regions show variation over 3 seeds.}
\label{fig:sp_closed_format_appendix}
\end{figure*}

\begin{figure*}[t]
\begin{center}
\includegraphics[width=0.7\linewidth]{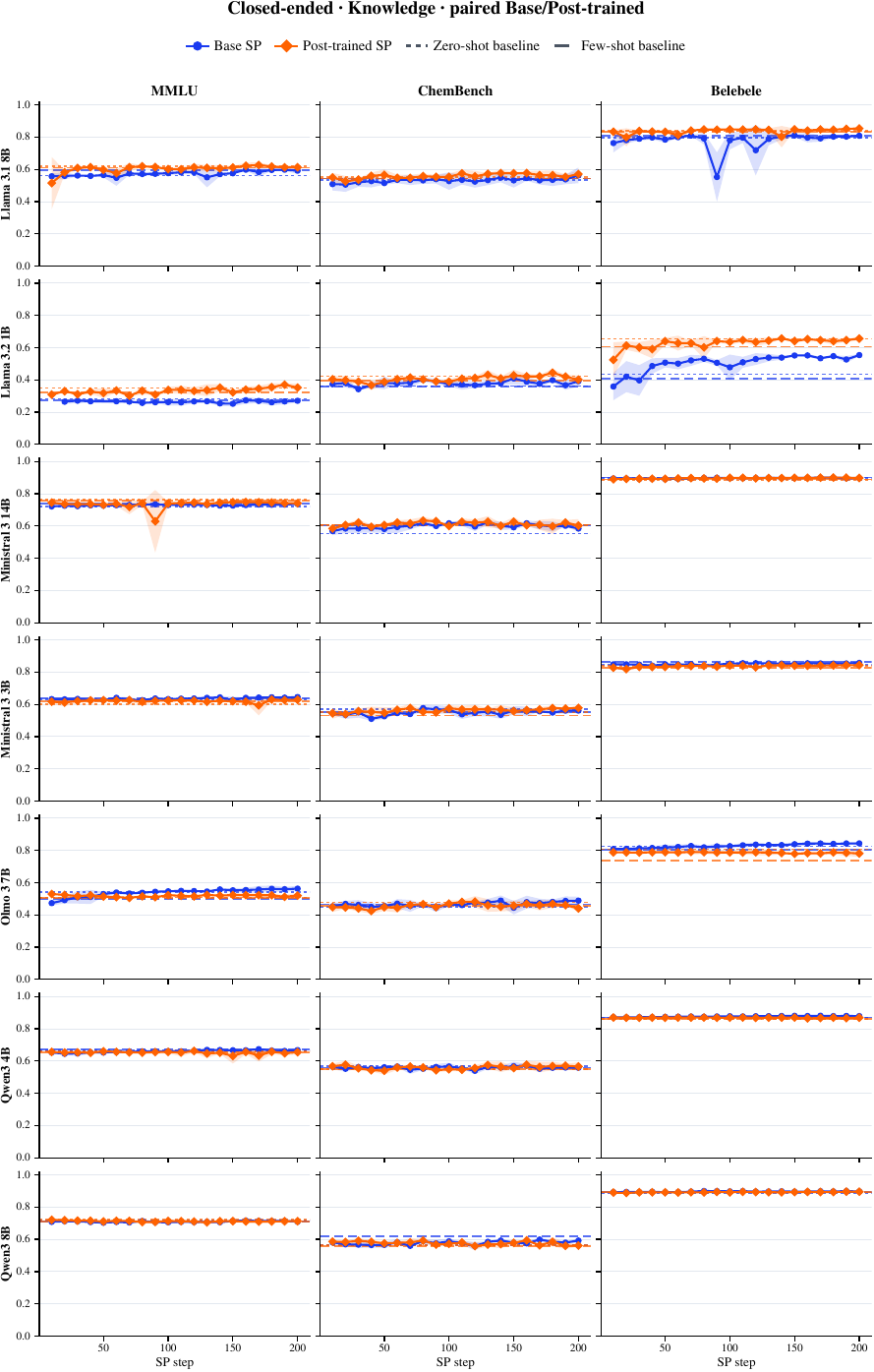}
\end{center}
\caption{\textbf{Soft prompt tuning does not add any substantial task knowledge to the model -- closed ended datasets.} We show that in nearly all of the settings soft-prompt tuned model stays aligned with the baseline knowledge accuracy throughout the training. Only the Llama3.2-1B model on Belebele dataset shows a slight increase. Shaded regions show variation over 3 seeds.}
\label{fig:sp_closed_knowledge_appendix}
\end{figure*}

In this section, we provide per dataset and model plots of knowledge and format-following accuracy on closed-ended datasets. Figure~\ref{fig:sp_closed_format_appendix} represents the format-following plots and Figure~\ref{fig:sp_closed_knowledge_appendix} represents the knowledge accuracy ones.

\subsection{Open-Ended Datasets Knowledge and Format-Following Accuracy Plots}
\label{appendix_open_ended_knowledge_format}

\begin{figure*}[t]
\begin{center}
\includegraphics[width=1\linewidth]{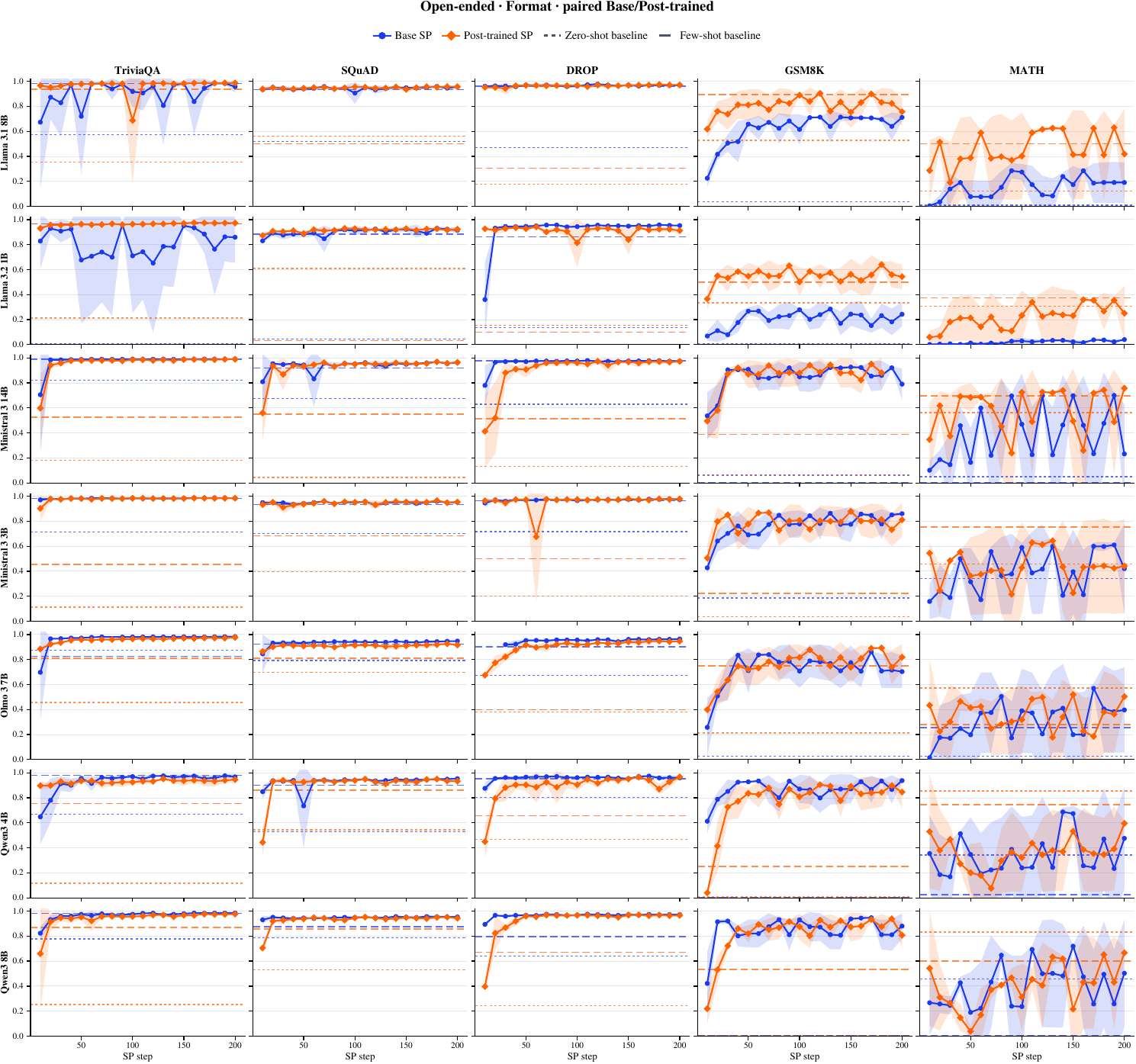}
\end{center}
\caption{\textbf{Soft prompt tuning maximizes format following -- open ended datasets.} Final format-following performance ranges from 80-100\% with the exception of the MATH dataset. We observed that this dataset is rather hard for maximizing format-following up to 100\% since it contains complex LaTeX format in the answers. It is also important to note that on math datasets (GSM8K and MATH) soft-prompt vectors are tuned on the reasoning traces as well, therefore it is harder to learn the overall format compared to other shorter datasets. Therefore, the fluctuation is expected.}
\label{fig:sp_open_format_appendix}
\end{figure*}

\begin{figure*}[t]
\begin{center}
\includegraphics[width=1\linewidth]{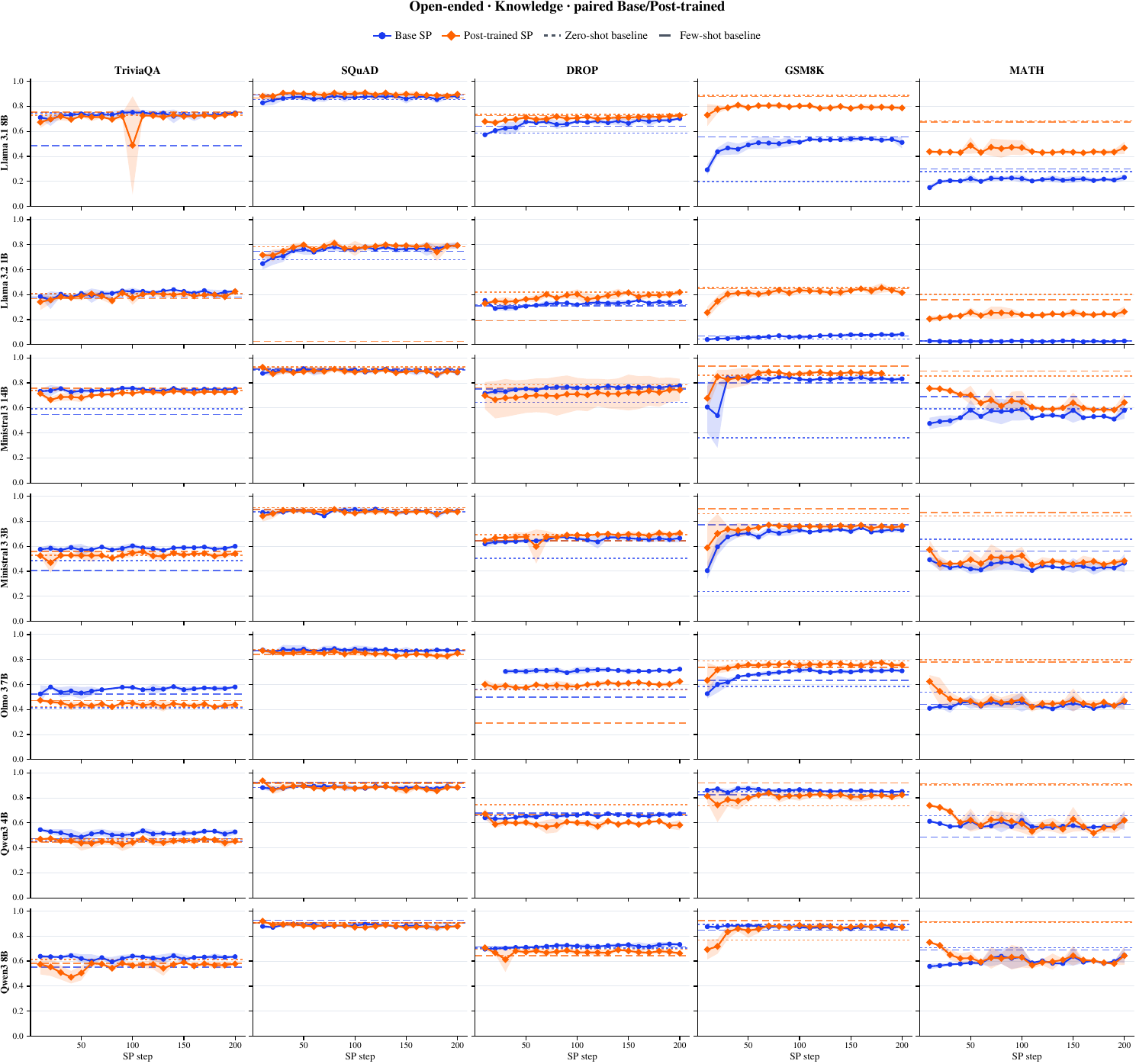}
\end{center}
\caption{\textbf{Soft prompt tuning does not add any systematic additional task knowledge to the model -- open ended datasets.} We find that, in nearly all settings, soft-prompt-tuned models remain aligned with the zero- or few-shot baseline knowledge accuracy throughout training. Only 4 out of 35 cases show slight fluctuations in knowledge accuracy. Given the limitations of LLM-judge-based knowledge evaluation (\ref{llm_judge_failure_examples}), and the fact that similar fluctuations also occur between zero- and few-shot baselines, we do not interpret these changes as meaningful knowledge gain.}
\label{fig:sp_open_knowledge_appendix}
\end{figure*}

In this section, we provide per dataset and model plots of knowledge and format-following accuracy on open-ended datasets. Figure~\ref{fig:sp_open_format_appendix} represents the format-following plots and Figure~\ref{fig:sp_open_knowledge_appendix} represents the knowledge accuracy ones.

\subsection{Long-context Datasets Format-Following Accuracy and Benchmark Accuracy Plots}
\label{appendix_long_context_knowledge_format}

\begin{figure*}[t]
\begin{center}
\includegraphics[width=0.75\linewidth]{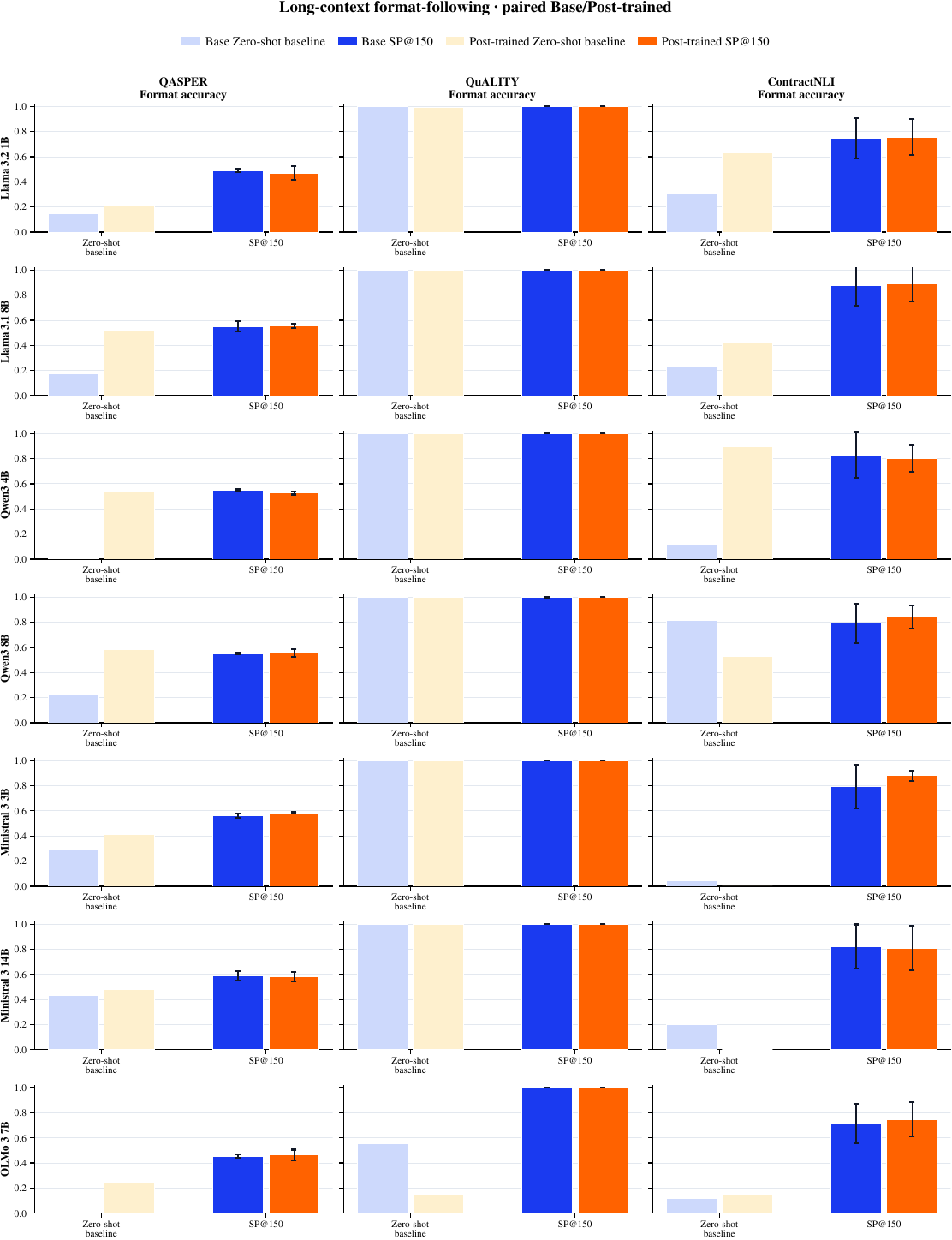}
\end{center}
\caption{\textbf{Soft-prompt tuning provides better format-following on long-context benchmarks.} Across all setups, soft-prompt tuning mostly achieves equal or better format-following scores than the zero-shot baseline. On ContractNLI and QuALITY, format-following reaches approximately 80–100\% the same as short-context open-ended datasets, whereas QASPER achieves 50–60\%. This difference is partly attributable to task structure: QuALITY requires a single answer choice (A–D), and ContractNLI uses one of three fixed labels, while QASPER requires variable-length free-text answers or \textit{Unanswerable}, making output structure and stopping behavior more difficult to learn. Moreover, the LLM-based format judge may partially conflate correctness with format compliance, where validly formatted but incorrect or differently phrased answers can be classified as format failures. Despite these challenges, soft-prompt tuning consistently improves over the QASPER baseline. Longer training or a multi-judge setup may further improve QASPER format compliance and provide more accurate format-following results. We identify this as a limitation and direction for future work. Error bars show variation across soft-prompt seeds.}
\label{fig:sp_long_context_format_appendix}
\end{figure*}

\begin{figure*}[t]
\begin{center}
\includegraphics[width=1\linewidth]{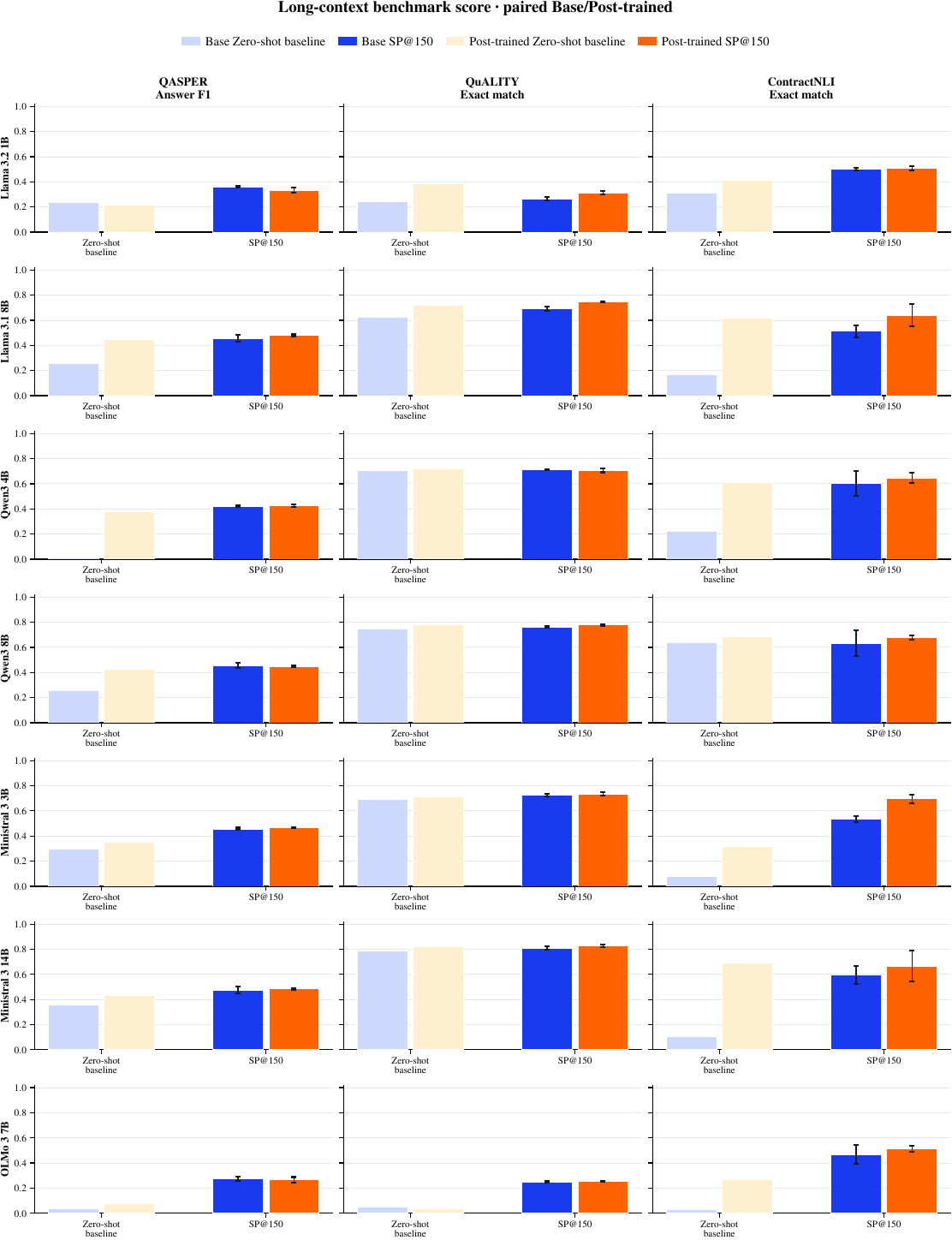}
\end{center}
\caption{\textbf{Soft-prompt tuning generally improves performance on long-context benchmarks, with the largest gains for base models.} At step 150, soft prompts outperform the zero-shot baseline for nearly all model pairs on QASPER and ContractNLI, with particularly large improvements for base models and OLMo~3~7B. Improvements on QuALITY are smaller because most models already achieve strong zero-shot scores, leaving less headroom. A few post-trained models show small decreases, but these do not alter the overall trend. QASPER is evaluated using answer F1, while QuALITY and ContractNLI use exact match error bars show variation across soft-prompt seeds.}
\label{fig:sp_long_context_task_appendix}
\end{figure*}

In this section, we provide per dataset and model plots of format-following accuracy and benchmark scores on open-ended long-context datasets. Figure~\ref{fig:sp_long_context_format_appendix} represents the format-following plots and Figure~\ref{fig:sp_long_context_task_appendix} represents the benchmark scores.

\subsection{Open-ended Dataset Benchmark Accuracy Plots}
\label{appendix_benchmark_acc}

\begin{figure*}[t]
\begin{center}
\includegraphics[width=1\linewidth]{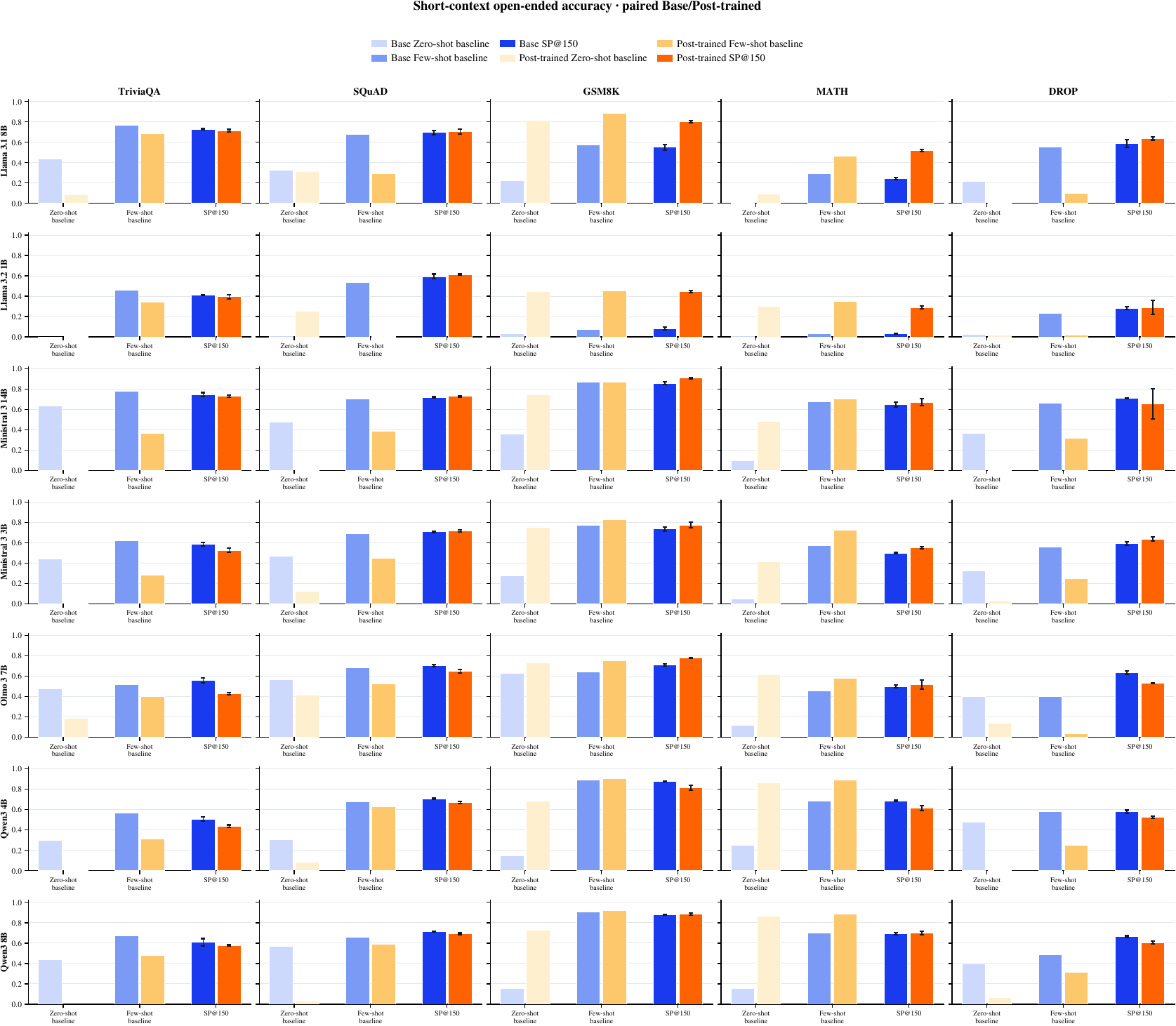}
\end{center}
\caption{\textbf{Soft-prompt tuning brings base and post-trained models closer together on open-ended benchmarks and soft-prompt tuned base models scores align well with the well-established baselines.} Under soft-prompt tuning, base models align well with the few-shot baseline scores on the  well-engineered benchmarks, whereas instruction-tuned models show better signal with soft-prompt tuning. Further, soft-prompt tuning brings the performances of base and instruction-tuned models close together unlike zero- and few-shot prompting, consistent with findings that suitably prompted base models can match instruction-tuned ones \citep{munjal_instruction-tuned_2026}.}

\label{fig:sp_baseline_per_appendix}
\end{figure*}

Figure~\ref{fig:sp_baseline_per_appendix} shows the per dataset and model open-ended benchmark accuracy scores for base and post-trained models under zero-shot, few-shot prompting and soft-prompt tuning. Soft-prompt results are taken at checkpoint 150 and averaged over 3 seeds.

\subsection{Olmo-2-32B Judge - Exact Match Alignment}
\label{appendix_olmo_alignment}
\begin{figure*}[t]
\begin{center}
\includegraphics[width=0.7\linewidth]{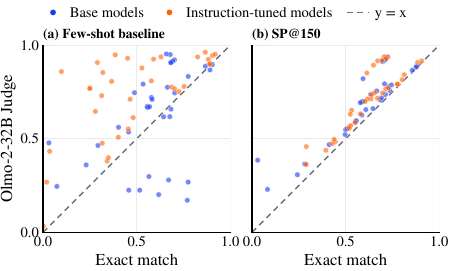}
\end{center}
\caption{\textbf{Soft-prompt tuning provides better judge-exact match metric alignment.} -- \textit{Open-ended datasets}. Consistent with the Qwen-3-32B results, soft-prompt tuning increases alignment between exact match and Olmo-2-32B compared to few-shot baseline. Soft-prompt tuning achieves this by letting models commit to one clear answer, whereas few-shot baseline often requires the judge to locate the answer within noisy text.}
\label{fig:olmo_alignment_appendix}
\end{figure*}

In this section, we show the alignment between Olmo-2-32B judge and exact match metric across few-shot and soft-prompt tuned model answers. Fig.~\ref{fig:olmo_alignment_appendix} shows how soft-prompt tuning increases the alignment between judge scores and exact match scores, supporting the main paper claims on Qwen-3-32B judge (Fig.~\ref{fig:llm_judge_alignment}).

\subsection{Ablating the Number of Soft-Prompt Vectors}
\label{softprompt_vector_ablation}

\begin{figure*}[t]
\begin{center}
\includegraphics[width=1\linewidth]{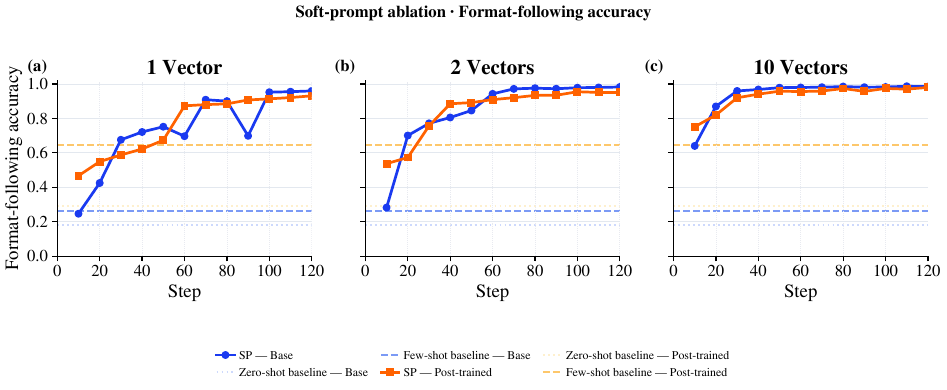}
\end{center}
\caption{\textbf{Lowering the number of soft-prompt vectors slows down format adherence}. Format-following accuracy tested on Belebele and Chembench datasets, Qwen3-4B and Olmo3-7B models. Results are aggregated over these 2 models and 2 datasets. Since 10 soft-prompt vectors provide much faster convergence while keeping the compute requirements low, we chose to use 10 soft-prompt vectors in our experiments.}
\label{fig:sp_vector_ablation}
\end{figure*}

In this section, we show the format-following accuracy curves for different numbers of soft-prompt vectors (1, 2, and 10). We conduct this ablation to see the effect of changing the number of soft-prompt vectors on format-following behavior. Each plot, in the Fig.~\ref{fig:sp_vector_ablation} shows results that are aggregated over 2 models (Qwen3-4B, Olmo3-7B) and 2 datasets (Belebele, Chembench). We see that, as expected, lowering number of soft-prompt vectors penalizes the format-following curves and leads to a much slower learning. Therefore, we chose to use 10 soft-prompt vectors in our experiments.

\clearpage
\subsection{Prompts for Benchmarks}
\label{appendix_benchmark_prompts}

Below we list the few-shot and soft-prompt tuning prompt variants for each closed- and open-ended dataset. %

\subsubsection{Open-ended Benchmarks}
The few-shot baselines use different number of fixed demonstrations that are taken from evaluation-harness implementations for each benchmark. SP prepends the ten learned embeddings before the prompt start. For the zero-shot baselines, we use the few-shot prompts with the demonstrations removed. %

\begin{promptbox}[TriviaQA: 5-shot]\footnotesize
    \textbf{Question:} From which country did Angola achieve independence in 1975?
    
    \textbf{Answer:} Portugal
    
    \(\ldots\) \textit{(4 further fixed demonstrations)}
    
    \textbf{Question:} \textit{\{question\}}
    
    \textbf{Answer:}
\end{promptbox}
    
\begin{promptbox}[TriviaQA: SP]\footnotesize
    \(\langle\mathrm{SP}_1\rangle \cdots
    \langle\mathrm{SP}_{10}\rangle\)
    
    Here is a trivia question for you. 
    
    \textbf{Question:} \textit{\{question\}}
    
    \textbf{Answer:}
\end{promptbox}

\begin{promptbox}[SQuAD: 5-shot]\footnotesize
    The following are reading comprehension questions, where the answer to each
    question is a segment of text from the corresponding background text.
    
    \textbf{Title:} University\_of\_Notre\_Dame
    
    \textbf{Background:} Architecturally, the school has a Catholic character. Atop the
    Main Building's gold dome is a golden statue of the Virgin Mary. Immediately
    in front of the Main Building and facing it, is a copper statue of Christ
    with arms upraised with the legend ``Venite Ad Me Omnes''. Next to the Main
    Building is the Basilica of the Sacred Heart. Immediately behind the
    basilica is the Grotto, a Marian place of prayer and reflection. It is a
    replica of the grotto at Lourdes, France where the Virgin Mary reputedly
    appeared to Saint Bernadette Soubirous in 1858. At the end of the main drive
    (and in a direct line that connects through 3 statues and the Gold Dome), is
    a simple, modern stone statue of Mary.
    
    \textbf{Question:} To whom did the Virgin Mary allegedly appear in 1858 in Lourdes
    France?
    
    \textbf{Answer:} Saint Bernadette Soubirous
    
    \(\ldots\) \textit{(4 further fixed demonstrations)}
    
    \textbf{Title:} \textit{\{title\}}
    
    \textbf{Background:} \textit{\{context\}}
    
    \textbf{Question:} \textit{\{question\}}
    
    \textbf{Answer:}
\end{promptbox}
    
\begin{promptbox}[SQuAD: SP]\footnotesize
    \(\langle\mathrm{SP}_1\rangle \cdots
    \langle\mathrm{SP}_{10}\rangle\)
    
    Answer the question based on the given context.
    
    \textbf{Context:} \textit{\{context\}}
    
    \textbf{Question:} \textit{\{question\}}
    
    \textbf{Answer:}
    \end{promptbox}
    
    \begin{promptbox}[DROP: 5-shot]\footnotesize
    The following are reading comprehension questions, where the answer to each
    question is either a segment of text from the corresponding passage, a
    number, or a date (containing any of the date, month, and/or year
    components). Some questions may require you to pull together information
    pieces from the passage and reason over them.
    
    \textbf{Passage:} To start the season, the Lions traveled south to Tampa, Florida to
    take on the Tampa Bay Buccaneers. The Lions scored first in the first quarter
    with a 23-yard field goal by Jason Hanson. The Buccaneers tied it up with a
    38-yard field goal by Connor Barth, then took the lead when Aqib Talib
    intercepted a pass from Matthew Stafford and ran it in 28 yards. The Lions
    responded with a 28-yard field goal. In the second quarter, Detroit took the
    lead with a 36-yard touchdown catch by Calvin Johnson, and later added more
    points when Tony Scheffler caught an 11-yard TD pass. Tampa Bay responded
    with a 31-yard field goal just before halftime. The second half was
    relatively quiet, with each team only scoring one touchdown. First,
    Detroit's Calvin Johnson caught a 1-yard pass in the third quarter. The
    game's final points came when Mike Williams of Tampa Bay caught a 5-yard
    pass.  The Lions won their regular season opener for the first time since
    2007
    
    \textbf{Question:} How many field goals did the Lions score?
    
    \textbf{Answer:} 2
    
    \(\ldots\) \textit{(4 further fixed demonstrations)}
    
    \textbf{Passage:}  \textit{\{passage\}}
    
    \textbf{Question:} \textit{\{question\}}
    
    \textbf{Answer:}
\end{promptbox}
    
\begin{promptbox}[DROP: SP]\footnotesize
    \(\langle\mathrm{SP}_1\rangle \cdots
    \langle\mathrm{SP}_{10}\rangle\)
    
    \textbf{Passage:}  \textit{\{passage\}}
    
    \textbf{Question:} \textit{\{question\}}
    
    \textbf{Answer:}
    \end{promptbox}
    
    \begin{promptbox}[GSM8K: 8-shot]\footnotesize
    \textbf{Question:} There are 15 trees in the grove. Grove workers will plant trees in
    the grove today. After they are done, there will be 21 trees. How many trees
    did the grove workers plant today?
    
    \textbf{Answer:} There are 15 trees originally. Then there were 21 trees after some
    more were planted. So there must have been \(21 - 15 = 6\). So the answer is
    6.
    
    \(\ldots\) \textit{(7 further fixed demonstrations)}
    
    \textbf{Question:} \textit{\{question\}}
    
    \textbf{Answer:}
    \end{promptbox}
    
    \begin{promptbox}[GSM8K: SP]\footnotesize
    \(\langle\mathrm{SP}_1\rangle \cdots
    \langle\mathrm{SP}_{10}\rangle\)
    
    \textbf{Question:} \textit{\{question\}}
    
    \textbf{Answer:}
\end{promptbox}
    
\begin{promptbox}[MATH: 4-shot]\footnotesize
    \textbf{Problem:}
    
    Find the domain of the expression
    \(\frac{\sqrt{x-2}}{\sqrt{5-x}}\).\}
    
    \textbf{Solution:} The expressions inside each square root must be non-negative.
    Therefore, \(x-2 \ge 0\), so \(x\ge2\), and \(5 - x \ge 0\), so \(x \le 5\).
    Also, the denominator cannot be equal to zero, so \(5-x>0\), which gives
    \(x<5\). Therefore, the domain of the expression is
    \( boxed([2,5)) \).
    
    \textbf{Final Answer:} The final answer is \([2,5)\). I hope it is correct.
    
    \(\ldots\) \textit{(3 further fixed demonstrations)}
    
    \textbf{Problem:}
    
    \textit{\{problem\}}
    
    \textbf{Solution:}
    \end{promptbox}
    
    \begin{promptbox}[MATH: SP]\footnotesize
    \(\langle\mathrm{SP}_1\rangle \cdots
    \langle\mathrm{SP}_{10}\rangle\)
    
    Here is a question for you.
    
    \textit{\{problem\}}
    
    \textbf{Solution:}
\end{promptbox}

\subsubsection{Closed-ended benchmarks}
    
The few-shot baselines use three random demonstrations for each closed-ended benchmark. SP uses the corresponding zero-shot textual template and prepends the ten learned embeddings. Candidate answers are scored by log-likelihood.

\begin{promptbox}[MMLU: 3-shot]\footnotesize
    The following are multiple choice questions (with answers) about
    \textit{\{subject\}}.

    Answer with the full text of the correct answer.

    \textbf{Here is an example:}

    \textbf{Question:} Which device converts kinetic energy into electrical energy?

    - calculator

    - generator

    - electric motor

    - electromagnet

    \textbf{Answer:} - generator
    
    \(\ldots\) \textit{(2 other random demonstrations; examples vary by item)}

    \textbf{Here is the final question:}

    \textbf{Question:} \textit{\{question\}}

    - \textit{\{choice 1\}}

    - \textit{\{choice 2\}}

    \(\vdots\)

    \textbf{Answer:}
\end{promptbox}
    
\begin{promptbox}[MMLU: SP]\footnotesize
    \(\langle\mathrm{SP}_1\rangle \cdots
    \langle\mathrm{SP}_{10}\rangle\)
    
    The following are multiple choice questions (with answers) about
    \textit{\{subject\}}.
    
    Answer with the full text of the correct answer.
    
    \textbf{Question:} \textit{\{question\}}
    
    - \textit{\{choice 1\}}
    
    - \textit{\{choice 2\}}
    
    \(\vdots\)
    
    \textbf{Answer:}
\end{promptbox}
    
\begin{promptbox}[ChemBench: 3-shot]\footnotesize
    The following is a question about chemistry. 
    
    Answer with the full text of the correct answer.
    
    \textbf{Here are some examples:}
    
    \textbf{Question:} Which is more toxic by inhalation? Hydrogen chloride or phosgene?
    
    A. Hydrogen chloride, because hydrogen chloride dissolves well in the water
    of the mucous membranes
    
    B. Hydrogen chloride, because hydrogen chloride has a direct corrosive
    effect
    
    C. Phosgene, because phosgene is colorless and produces genotoxic
    metabolites after metabolic activation by CYP2E1
    
    D. Phosgene, because phosgene is less soluble in water than hydrogen
    chloride and can therefore penetrate into the alveoli
    
    \textbf{Answer:} D. Phosgene, because phosgene is less soluble in water than hydrogen
    chloride and can therefore penetrate into the alveoli
    
    \(\ldots\) \textit{(2 further random demonstrations)}
    
    \textbf{Here is the final question:}
    
    \textbf{Question:} \textit{\{question\}}
    
    A. \textit{\{choice A\}}
    
    B. \textit{\{choice B\}}
    
    \(\vdots\)
    
    \textbf{Answer:}
\end{promptbox}
    
\begin{promptbox}[ChemBench: SP]\footnotesize
    \(\langle\mathrm{SP}_1\rangle \cdots
    \langle\mathrm{SP}_{10}\rangle\)
    
    The following is a question about chemistry. 
    
    Answer with the full text of the correct answer.
    
    \textbf{Question:} \textit{\{question\}}
    
    A. \textit{\{choice A\}}
    
    B. \textit{\{choice B\}}
    
    \(\vdots\)
    
    \textbf{Answer:}
\end{promptbox}
    
\begin{promptbox}[Belebele: 3-shot]\footnotesize
    You are solving Belebele multiple-choice reading comprehension.
    
    Answer with the full text of the correct answer.
    
    \textbf{Here are some examples:}
    
    Heat capacity is the amount of internal energy that the substance can store.
    A large heat capacitance means the substance can store a lot of internal
    energy and thus the temperature changes slowly. Aluminum foil has a small
    heat capacitance and water has a large one.
    
    \textbf{Question:} What is the term for the amount of internal energy that the
    substance can store?
    
    A. mass capacity
    
    B. temperature capacity
    
    C. humidity capacity
    
    D. heat capacity
    
    \textbf{Answer:} D. heat capacity
    
    \(\ldots\) \textit{(2 further random demonstrations)}
    
    \textbf{Here is the final question:}
    
    \textit{\{passage\}}
    
    \textbf{Question:} \textit{\{question\}}
    
    A. \textit{\{choice A\}}
    
    B. \textit{\{choice B\}}
    
    C. \textit{\{choice C\}}
    
    D. \textit{\{choice D\}}
    
    \textbf{Answer:}
\end{promptbox}
    
\begin{promptbox}[Belebele: SP]\footnotesize
    \(\langle\mathrm{SP}_1\rangle \cdots
    \langle\mathrm{SP}_{10}\rangle\)
    
    You are solving Belebele multiple-choice reading comprehension.
    
    Answer with the full text of the correct answer.
    
    \textit{\{passage\}}
    
    \textbf{Question:} \textit{\{question\}}
    
    A. \textit{\{choice A\}}
    
    B. \textit{\{choice B\}}
    
    C. \textit{\{choice C\}}
    
    D. \textit{\{choice D\}}
    
    \textbf{Answer:}
\end{promptbox}

\subsubsection{Long-context benchmarks}
\label{long_context_prompts}
Our long-context experiments have only the zero-shot prompting variants since few-shot baselines do not exist for these benchmarks.
    
\begin{promptbox}[Qasper: zero-shot]\footnotesize
    You are given a scientific article and a question. Answer the question as
    concisely as you can, using a single phrase or sentence if possible. If the
    question cannot be answered based on the information in the article, write
    ``unanswerable''. If the question is a yes/no question, answer ``yes'',
    ``no'', or ``unanswerable''.
    
    \textbf{Article:}
    
    \textit{\{article\}}
    
    \textbf{Question:}
    
    \textit{\{question\}}
    
    \textbf{Answer:}
\end{promptbox}
    
\begin{promptbox}[Qasper: SP]\footnotesize
    \(\langle\mathrm{SP}_1\rangle \cdots
    \langle\mathrm{SP}_{10}\rangle\)
    
    \textbf{Article:}
    
    \textit{\{article\}}
    
    \textbf{Question:}
    
    \textit{\{question\}}
    
    \textbf{Answer:}
    \end{promptbox}
    
    \begin{promptbox}[QuALITY: zero-shot]\footnotesize
    \textbf{Story:}
    
    \textit{\{story\}}
    
    \textbf{Question and Possible Answers:}
    
    \textit{\{question\}}
    
     (A) \textit{\{option A\}}
    
     (B) \textit{\{option B\}}
    
     (C) \textit{\{option C\}}
    
     (D) \textit{\{option D\}}
    
    \textbf{Answer:}
\end{promptbox}
    
\begin{promptbox}[QuALITY: SP]\footnotesize
    \(\langle\mathrm{SP}_1\rangle \cdots
    \langle\mathrm{SP}_{10}\rangle\)
    
    \textbf{Story:}
    
    \textit{\{story\}}
    
    \textbf{Question and Possible Answers:}
    
    \textit{\{question\}}
    
     (A) \textit{\{option A\}}
    
     (B) \textit{\{option B\}}
    
     (C) \textit{\{option C\}}
    
     (D) \textit{\{option D\}}
    
    \textbf{Answer:}
\end{promptbox}
    
\begin{promptbox}[ContractNLI: zero-shot]\footnotesize
    Determine whether the hypothesis is entailed by, contradicted by, or not
    mentioned in the contract. Respond with exactly one label: Entailment,
    Contradiction, or Not mentioned.
    
    \textbf{Contract:}
    
    \textit{\{contract\}}
    
    \textbf{Hypothesis:}
    
    \textit{\{hypothesis\}}
    
    \textbf{Conclusion:}
\end{promptbox}
    
\begin{promptbox}[ContractNLI: SP]\footnotesize
    \(\langle\mathrm{SP}_1\rangle \cdots
    \langle\mathrm{SP}_{10}\rangle\)
    
    \textbf{Contract:}
    
    \textit{\{contract\}}
    
    \textbf{Hypothesis:}
    
    \textit{\{hypothesis\}}
    
    \textbf{Conclusion:}
\end{promptbox}

\subsection{LLM-as-a-judge Prompts}
\label{appendix_llm_judge}

In this section, we provide the specific prompt templates used for our LLM-as-a-judge evaluation of the model's knowledge and format-following. 

\subsubsection{Knowledge Accuracy Prompt}

These prompts are designed to assess knowledge accuracy by instructing a judge model to determine if a candidate completion is semantically equivalent to the ground-truth answers, regardless of formatting or additional irrelevant text.

\begin{figure*}[t]
\centering
\begin{promptbox}[Knowledge Judge Prompt]\footnotesize
    You are a judge that checks the correctness of answers.

        You will be given:
        \begin{itemize}
        \item A ground-truth answer (GT)
        \item A prediction produced by a model
        \end{itemize}

        Your task:
        \begin{itemize}
        \item Determine whether the prediction contains a correct answer to the original question.
        \item First, identify the final correct answer from the GT.
        \item Then, read the entire prediction and check whether it contains that answer.
        \item The prediction does not need to provide reasoning or explanations; correctness is the only criterion.
        \item The prediction can be in any form and may include additional or random generations (e.g., repeated questions, new questions, extra answers, irrelevant text). Do not penalize this and ignore it for correctness.
        \item The prediction does NOT need to match the GT in formatting or phrasing; only the underlying answer correctness matters.
        \item Ignore formatting, structure, language choice, explanation quality, and reasoning steps.
        \item You MUST choose exactly one: "1" (correct) or "0" (incorrect).

        Output format (STRICT):
        \item Output MUST be a single-line JSON object, and nothing else.
        \item Do NOT add any text before or after the JSON.
        \item The JSON MUST have exactly these keys:
        \item "accuracy": "1" or "0"
        \item "reason\_short": a short string (one or two sentences)
        
        \end{itemize}
        Example of a valid output:
        
        \{\{"accuracy": "1", "reason\_short": "<Brief explanation of why the answer is correct.>"\}\}
\end{promptbox}
\end{figure*}

\subsubsection{Format-Following Accuracy Prompt}

To measure format-following accuracy, we use a specialized judge prompt that ignores the factual correctness of the content and focuses exclusively on whether the model adhered to the specified output constraints. This allows us to quantify the "formatting bottleneck" independently of the model's underlying knowledge.

\begin{figure*}[t]
\centering
\begin{promptbox}[Format Following Judge Prompt (1/2)]\footnotesize

            You are a judge that evaluates FORMATTING STYLE ONLY.

            CRITICAL RULE — READ THIS CAREFULLY:

            You must COMPLETELY IGNORE the CONTENT of the final answer. Do NOT look at what number, expression, name, or value the prediction produces. The final answer could be completely WRONG — that is 100\% irrelevant. You are NOT a correctness checker. You are ONLY a format checker.

            If the prediction uses the right formatting conventions (e.g., LaTeX + boxed\{\{\}\}, or <<>> + \#\#\#\#, or a short phrase) but gets the WRONG answer, you MUST output "1".

            If you catch yourself thinking "the answer is incorrect" or "the answer differs from the GT" — STOP. That thought is irrelevant. Only formatting matters.

            You will be given:
            
            \begin{itemize}
                \item A ground-truth answer (GT) — use this ONLY as a reference for the expected formatting style
                \item A model prediction — evaluate whether its formatting matches the GT's style
            \end{itemize}
            
            Formatting criteria to check (in order of importance):

            1. STOPPING BEHAVIOR (most important):
            \begin{itemize}
               \item The prediction must produce exactly ONE answer and then STOP.
               \item If the prediction repeats the answer, re-generates the question, loops, or keeps writing after the final answer, that is a formatting failure → "0".
               \item Outputs that contain "Here is the answer to the question:" or echo the prompt after answering are badly formatted → "0".
               \item Outputs that get stuck in a generation loop (e.g., repeating the same calculation or text pattern over and over) are badly formatted → "0".
            \end{itemize}

            2. STRUCTURAL PATTERN — match the GT's style:
               The GT defines what "correct format" looks like. Different GTs have different styles. You must adapt to whatever style the GT uses:

               a) Step-by-step math with special notation (e.g., GSM8K style):
               \begin{itemize} 
                  \item GT uses inline calculations like <<expression=result>>result
                  \item GT ends with "\#\#\#\# <number>" as the final answer marker
                  \item Prediction must also use <<>> notation and \#\#\#\# ending
                  \item Different number of reasoning steps is fine
               \end{itemize}

               b) LaTeX/mathematical prose (e.g., MATH style):
               \begin{itemize}
                
                  \item GT uses LaTeX notation ($...$, \\boxed{{}}, \\begin{{align*}}, \\frac{{}}{{}}, etc.)
                  \item GT wraps the final answer in \\boxed{{}}
                  \item Prediction must also use LaTeX and \\boxed{{}} for the final answer
                  \item Shorter or longer derivations are fine
                  \item Using completely DIFFERENT mathematical methods, theorems, or reasoning approaches is fine (e.g., using partial derivatives instead of substitution, or a geometric argument instead of algebraic — these are NOT format issues)
                  \item The specific proof technique, solution strategy, or level of detail does NOT matter — only the surface-level formatting (LaTeX + \\boxed{{}}) matters
               \end{itemize}
\end{promptbox} 
\end{figure*}
\begin{figure*}
\centering
\begin{promptbox}[Format Following Judge Prompt (2/2)]\footnotesize
           c) Short factual answers (e.g., trivia style):
           \begin{itemize}
              \item GT is a brief answer (a name, a word, a number, a short phrase)
              \item Prediction should also be a brief answer
              \item Minor text differences are COMPLETELY FINE: abbreviations ("W.B." vs "William Butler"), punctuation ("A.A. Milne" vs "A A Milne"), capitalization ("The Firm" vs "THE FIRM"), quoting differences — all fine
              \item These are format-irrelevant surface variations → "1"
           \end{itemize}
            3. WHAT TO IGNORE — COMPLETELY IRRELEVANT (do NOT mention any of these in your reason):
            \begin{itemize}
               \item THE FINAL ANSWER VALUE: whether it is right, wrong, different, or matches the GT — NEVER consider this
               \item Whether the math/logic is correct or incorrect
               \item Whether intermediate calculations are right or wrong
               \item The reasoning method, proof technique, or solution strategy used (e.g., Law of Sines vs Law of Cosines, substitution vs partial derivatives, factorization vs completing the square — all irrelevant)
               \item Whether the derivation follows the same steps or uses the same theorems as the GT
               \item The level of detail or length of the derivation (shorter or longer is fine)
               \item Minor wording differences, paraphrasing, or different variable names
               \item Slightly more or fewer reasoning steps
               \item Abbreviations, punctuation, or capitalization differences in short answers
               \item Different but equivalent notations (e.g., 0.5 vs 1/2)
            \end{itemize}

            SELF-CHECK BEFORE ANSWERING:
            \begin{itemize}
            \item Did I base my score ONLY on formatting (notation style, stopping behavior, structural markers)?
            \item Did I accidentally penalize the prediction for having a wrong or different answer? If yes → fix it.
            \item A prediction with wrong math but correct format (LaTeX + \\boxed\{\{\}\}, or <<>> + \#\#\#\#, or short answer) = "1".
            \end{itemize}

            Output format (STRICT):
            \begin{itemize}
            \item Output MUST be a single-line JSON object, and nothing else.
            \item Do NOT add any text before or after the JSON.
            \item The JSON MUST have exactly these keys:
                \begin{itemize} 
                  \item "accuracy": "1" or "0"
                  \item "reason\_short": a short string (one or two sentences)
                \end{itemize}
            \end{itemize}

            Example input will look like this:
            GROUND\_TRUTH:
            <<<GT\_START>>>
            ground truth inserted here.
            <<<GT\_END>>>

            Prediction:
            <<<C1\_START>>>
            prediction inserted here.
            <<<C1\_END>>>

            Example of a valid output:
            \{\{"accuracy": "1", "reason\_short": "<Brief explanation of why formatting is correct.>"\}\}
\end{promptbox} 
\end{figure*}

\clearpage
\subsection{LLM-judge Failure Examples}
\label{llm_judge_failure_examples}
    
\begin{promptbox}[LLM-as-a-Judge Knowledge-Accuracy Failures]
\footnotesize
\textbf{Dataset: SQuAD} \\
\textbf{Question:} What was Oklahoma's high-school dropout rate in 2007? \\
\textbf{Ground truth:} 3.1\% \\
\textbf{Prediction:} 31 percent \\
\textbf{LLM-judge score:} 1 \qquad
\textbf{Human score:} 0 \\
\textbf{Explanation:} The judge incorrectly interprets 31 as 3.1, although the
prediction is wrong by a factor of ten.
\medskip\hrule\medskip
\textbf{Dataset: TriviaQA} \\
\textbf{Question:} The office of the President of which South American country
is La Casa Rosada? \\
\textbf{Ground truth:} ARGENTINA \\
\textbf{Prediction:} argentina \\
\textbf{LLM-judge score:} 0 \qquad
\textbf{Human score:} 1 \\
\textbf{Explanation:} The prediction exactly matches the answer apart from
capitalization, but the judge incorrectly marks it as wrong.
\medskip\hrule\medskip
\textbf{Dataset: TriviaQA} \\
\textbf{Question:} Where in a vertebrate would you find the protein myoglobin? \\
\textbf{Ground truth:} Muscle tissue \\
\textbf{Prediction:} in muscles \\
\textbf{LLM-judge score:} 0 \qquad
\textbf{Human score:} 1 \\
\textbf{Explanation:} The prediction is semantically equivalent to the
ground truth, but the judge fails to recognize the paraphrase.
\medskip\hrule\medskip
\textbf{Dataset: SQuAD} \\
\textbf{Question:} Which carnival takes place in Oruro, Bolivia? \\
\textbf{Ground truth:} La Diablada \\
\textbf{Prediction:} La Diablada Carnival \\
\textbf{LLM-judge score:} 0 \qquad
\textbf{Human score:} 1 \\
\textbf{Explanation:} The prediction contains the complete ground-truth answer
and only adds the relevant descriptor \emph{Carnival}, but the judge rejects it.
\medskip\hrule\medskip
\textbf{Dataset: MATH} \\
\textbf{Question:} The vertex of $y=2x^2-4x+4$ is $(m,n)$. What is $m$? \\
\textbf{Ground truth:} $1$ \\
\textbf{Prediction:} $2$ \\
\textbf{LLM-judge score:} 1 \qquad
\textbf{Human score:} 0 \\
\textbf{Explanation:} The model derives the vertex $(1,2)$ but confuses $m$
and $n$ and reports $m=2$. The judge follows the flawed reasoning instead of
comparing the final answer with the ground truth.
\end{promptbox}

\begin{promptbox}[LLM-as-a-Judge Format-Following Accuracy Failures]
\footnotesize

\textbf{Dataset: QASPER} \\

\textbf{Question:} On top of BERT, does the RNN layer or the transformer layer
work better? \\

\textbf{GT:} Transformer over BERT (ToBERT) \\

\textbf{Prediction:} transformer layer \\

\textbf{LLM-Judge Format Score:} 0 \qquad
\textbf{Human Format Score:} 1 \\

\textbf{Failure:} The judge penalizes reduced specificity despite the prediction following the required short-answer format.

\medskip\hrule\medskip

\textbf{Dataset: ContractNLI} \\

\textbf{Hypothesis:} The Receiving Party may retain some Confidential
Information after its return or destruction. \\

\textbf{GT:} Not mentioned \\

\textbf{Prediction:} Entailment \\

\textbf{LLM-Judge Format Score:} 0 \qquad
\textbf{Human Format Score:} 1 \\

\textbf{Failure:} \emph{Entailment} is a valid ContractNLI label, the judge
conflates semantic correctness with format compliance.

\medskip\hrule\medskip

\textbf{Dataset: QASPER} \\

\textbf{Question:} What does a node in the network approach represent? \\

\textbf{GT:} Unanswerable \\

\textbf{Prediction:} country \\

\textbf{LLM-Judge Format Score:} 0 \qquad
\textbf{Human Format Score:} 1 \\

\textbf{Failure:} The prediction is knowledge-wise incorrect but follows the
required short-answer format. The judge therefore conflates semantic
correctness with format compliance.

\medskip\hrule\medskip

\textbf{Dataset: DROP} \\

\textbf{Question:} Which occurred first: Danzig becoming the most privileged
city in the Commonwealth or the ruler of Poland promising to pay 220,000
zlotys? \\

\textbf{GT:} The ruler promising to pay 220,000 zlotys \\

\textbf{Prediction:} Danzig becoming the most privileged city in the
Commonwealth \\

\textbf{LLM-Judge Format Score:} 0 \qquad
\textbf{Human Format Score:} 1 \\

\textbf{Failure:} Although the prediction is factually incorrect, it follows
the required short-answer format. The judge conflates semantic correctness
with format compliance.

\end{promptbox}

\newpage

\subsection{Item-Level Disagreement Examples Between Few-shot Baseline and Soft-Prompt Tuning}
\label{app:few_shot_sp_reasoning_examples}

In this section, we show examples on which the few-shot baseline and soft-prompt tuning disagree. The examples come from the well-established baselines where soft-prompt tuning and few-shot baseline have 83\% item-level agreement  (Fig.~\ref{fig:base_sp_olmes_align}). Table~\ref{tab:few_shot_sp_disagreement_examples} shows the examples from short-answer open-ended datasets like TriviaQA and SQUAD, whereas Table~\ref{tab:few_shot_sp_gsm8k_examples} shows examples from math datasets which require reasoning before answering.

\begin{table*}[t]
\centering
\footnotesize
\setlength{\tabcolsep}{4.5pt}
\renewcommand{\arraystretch}{1.18}

\begin{tabularx}{\textwidth}{
@{}
>{\raggedright\arraybackslash}p{2.25cm}
>{\raggedright\arraybackslash}X
>{\raggedright\arraybackslash}X
>{\raggedright\arraybackslash}X
@{}}
\toprule
\textbf{Failure type}
& \textbf{Question / gold answer}
& \textbf{Few-shot Baseline}
& \textbf{SP@150} \\
\midrule

\multicolumn{4}{@{}l}{\textbf{Few-shot Baseline wrong, SP right}} \\
\addlinespace[2pt]

Factual Failure
& Which Shakespeare play contains ``A plague on both your houses''? \newline
  \textit{Gold:} \texttt{Romeo and Juliet}
& \texttt{The Merchant of Venice}
& \textbf{\texttt{Romeo and Juliet}} \\

Factual Failure
& In which year was \textit{The Archers} first broadcast nationally? \newline
  \textit{Gold:} \texttt{1951}
& \texttt{1950}
& \textbf{\texttt{1951}} \\

Format Failure
& A Varsovian is a native of which capital? \newline
  \textit{Gold:} \texttt{Warsaw}
& \texttt{Warsaw Question: What is the name of the 2008 film\ldots}
& \textbf{\texttt{Warsaw}} \\

Format Failure
& DROP percentage question \newline
  \textit{Gold:} \texttt{99.26}
& \texttt{99.26\%}
& \textbf{\texttt{99.26}} \\

\midrule
\multicolumn{4}{@{}l}{\textbf{SP wrong, Few-shot Baseline right}} \\
\addlinespace[2pt]

Factual Failure
& Who was the original drummer and singer of the Monkees? \newline
  \textit{Gold:} \texttt{Micky Dolenz}
& \textbf{\texttt{Micky Dolenz}}
& \texttt{Davy Jones} \\

Factual Failure
& In which country is the southern Pillar of Hercules? \newline
  \textit{Gold:} \texttt{Morocco}
& \textbf{\texttt{Morocco}}
& \texttt{Spain} \\

Format Failure
& Who led Pontus in 85 BC? \newline
  \textit{Gold:} \texttt{Mithridates the Great}
& \textbf{\texttt{Mithridates the Great}}
& \texttt{Mithridates} \\

Format Failure
& Which queen is buried at Sudeley Castle? \newline
  \textit{Gold:} \texttt{Catherine Parr}
& \textbf{\texttt{Katherine Parr}}
& \texttt{Katherine Parr, the last wife of Henry VIII.} \\

\bottomrule
\end{tabularx}
\caption{Examples of item-level disagreement between few-shot baseline and soft-prompt tuning on short-answer open-ended datasets (TriviaQA, SQUAD, DROP)
}
\label{tab:few_shot_sp_disagreement_examples}
\end{table*}

\begin{table*}[t]
\centering
\small
\setlength{\tabcolsep}{6pt}
\renewcommand{\arraystretch}{1.18}

\providecommand{\correct}[1]{\textbf{\texttt{#1}}}

\begin{tabularx}{\textwidth}{
    @{}
    p{2.45cm}
    >{\raggedright\arraybackslash}X
    >{\raggedright\arraybackslash}X
    @{}
}
\toprule
\textbf{Failure type}
& \textbf{Few-shot Baseline Reasoning}
& \textbf{SP@150 Reasoning} \\
\midrule

\multicolumn{3}{@{}l}{\textbf{Few-shot Baseline wrong, SP right}} \\
\addlinespace[2pt]

Report Intermediate Answer
&
\multicolumn{2}{@{}p{0.78\textwidth}@{}}{
\textit{Question (GSM8K):} On Wednesday, 37 students played
kickball. On Thursday, 9 fewer students played. How many played
across both days?
\quad \textit{Gold:} \texttt{65}.
} \\
\addlinespace[1pt]
&
On Thursday, $37-9=28$ students played.
Therefore, the answer is \texttt{28}.
&
On Thursday, $37-9=28$ students played.
Across both days, $37+28=65$ students played.
Therefore, the answer is \correct{65}.
\\

\addlinespace[5pt]

Report Intermediate Answer
&
\multicolumn{2}{@{}p{0.78\textwidth}@{}}{
\textit{Question (MATH):} Two consecutive perfect squares
differ by 35. What is the greater square?
\quad \textit{Gold:}\texttt{324}.
} \\
\addlinespace[1pt]
&
Let the squares be $n^2$ and $(n+1)^2$.
Then $2n+1=35$, so $n=17$.
Therefore, the answer is \texttt{17}.
&
From $2n+1=35$, we obtain $n=17$.
The greater square is $(n+1)^2=18^2=324$.
Therefore, the answer is \correct{324}.
\\

\addlinespace[5pt]

Reasoning Mistake
&
\multicolumn{2}{@{}p{0.78\textwidth}@{}}{
\textit{Question (GSM8K):} Marina has 4.5 pounds of fudge.
Lazlo has 6 ounces less than 4 pounds. How many more ounces does
Marina have?
\quad \textit{Gold:} \texttt{14}.
} \\
\addlinespace[1pt]
&
Lazlo has $4$ pounds $-6$ ounces $=26$ ounces.
Marina has $4.5$ pounds $=90$ ounces.
Thus, Marina has $90-26=64$ more ounces.
The answer is \texttt{64}.
&
Marina has $4.5\times16=72$ ounces.
Lazlo has $4\times16-6=58$ ounces.
Thus, Marina has $72-58=14$ more ounces.
The answer is \correct{14}.
\\

\midrule

\multicolumn{3}{@{}l}{\textbf{Few-shot Baseline right, SP wrong}} \\
\addlinespace[2pt]

Report Intermediate Answer
&
\multicolumn{2}{@{}p{0.78\textwidth}@{}}{
\textit{Question (GSM8K):} Sally sold 20 cups of lemonade last
week and 30\% more this week. How many cups did she sell across
both weeks?
\quad \textit{Gold:} \texttt{46}.
} \\
\addlinespace[1pt]
&
This week, Sally sold $20+0.30\times20=26$ cups.
Across both weeks, she sold $20+26=46$ cups.
Therefore, the answer is \correct{46}.
&
Thirty percent of 20 is 6, so Sally sold
$20+6=26$ cups this week.
Therefore, the answer is \texttt{26}.
\\

\addlinespace[5pt]

Reasoning Mistake
&
\multicolumn{2}{@{}p{0.78\textwidth}@{}}{
\textit{Question (MATH):} How many units long is the segment
with endpoints $(1,2)$ and $(-4,-10)$?
\quad \textit{Gold:} \texttt{13}.
} \\
\addlinespace[1pt]
&
The distance is
$\sqrt{(-4-1)^2+(-10-2)^2}
 =\sqrt{25+144}
 =\sqrt{169}=13$.
Therefore, the answer is \correct{13}.
&
The squared differences sum to
$(-5)^2+(-12)^2=25+144=169$.
Therefore, the answer is \texttt{169}.
\\

\addlinespace[5pt]

State the Final Answer Wrong
&
\multicolumn{2}{@{}p{0.78\textwidth}@{}}{
\textit{Question (GSM8K):} A box of cereal originally costs
\$104 and is discounted by \$24. How much do 20 boxes cost?
\quad \textit{Gold:} \texttt{1600}.
} \\
\addlinespace[1pt]
&
The discounted price is $104-24=80$ dollars.
For 20 boxes, the total is $20\times80=1600$.
Therefore, the answer is \correct{1600}.
&
The discounted price is $104-24=80$ dollars.
For 20 boxes, the total is $20\times80=1600$.
Therefore, the answer is \texttt{160}.
\\

\bottomrule
\end{tabularx}

\caption{Examples of item-level disagreement between few-shot baseline and soft-prompt tuning on open-ended math datasets (GSM8K and MATH)
}
\label{tab:few_shot_sp_gsm8k_examples}
\end{table*}

\subsection{LoRA Experiment Results}
\label{lora_experiments_appendix}

In this section, we show our results on LoRA experiments. We conducted our LoRA experiments on the datasets Belebele, Chembench, MMLU, GSM8K, SQUAD and TriviaQA and we used the same models as in our soft-prompt tuning experiments. We used rank 8 for these experiments and trained all models for 3 different seeds. For LoRA we chose step 95 as the optimum step to maximize format-following without introducing task-specific knowledge. Although LoRA is trained for fewer steps (95 versus 150 on soft-prompt tuning), it inherently requires significantly more parameters compared to soft-prompts. With LoRA we train 0.27\% of a 7B model's parameters whereas with soft-prompts we only train 0.0006\%, making soft prompting significantly more parameter-efficient.

\begin{figure*}[t]
\centering
\begin{minipage}[t]{0.45\linewidth}
\makebox[\linewidth][l]{\textnormal{(a)}}\\[-2pt]
\includegraphics[width=\linewidth,trim=0 0 0 25,clip]{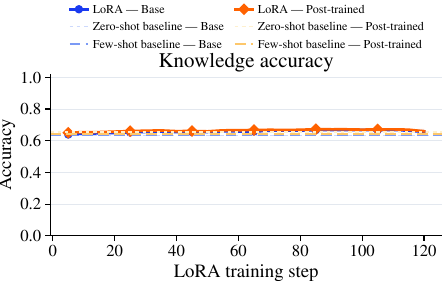}
\end{minipage}
\hfill
\begin{minipage}[t]{0.45\linewidth}
\makebox[\linewidth][l]{\textnormal{(b)}}\\[-2pt]
\includegraphics[width=\linewidth,trim=0 0 0 25,clip]{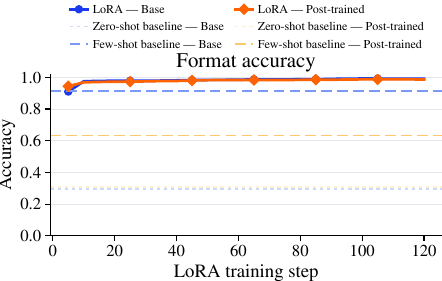}
\end{minipage}
\vspace{2pt}
\includegraphics[width=0.5\textwidth]{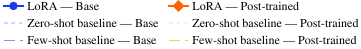}
\caption{\textbf{LoRA maximizes format-following without introducing additional knowledge.} -- \textit{closed-ended datasets.} (a) Knowledge and (b) format-following metrics across 120 tuning steps. All results are aggregated over 3 closed-ended datasets, 7 models, and 3 seeds.}
\label{fig:lora_close_format_knowledge_appendix}
\end{figure*}

\begin{figure*}[t]
\centering

\begin{minipage}[t]{0.4\linewidth}
\makebox[\linewidth][l]{\textnormal{(a)}}\\[-2pt]
\includegraphics[width=\linewidth,trim=0 0 0 25,clip]{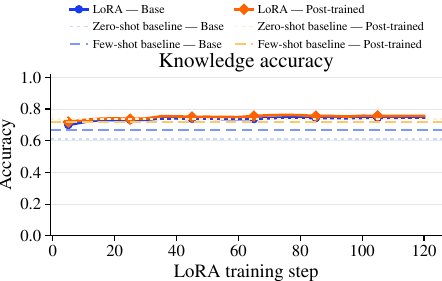}
\end{minipage}
\hfill
\begin{minipage}[t]{0.4\linewidth}
\makebox[\linewidth][l]{\textnormal{(b)}}\\[-2pt]
\includegraphics[width=\linewidth,trim=0 0 0 25,clip]{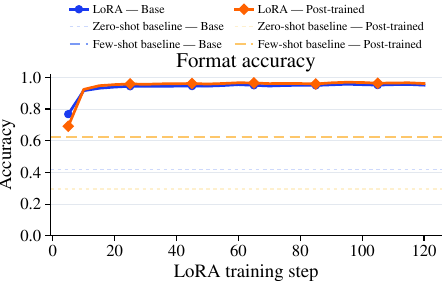}
\end{minipage}
\vspace{2pt}
\includegraphics[width=0.5\textwidth]{figures/appendix/lora/legend_lora_knowledge_format.pdf}
\caption{\textbf{LoRA maximizes format-following without introducing additional knowledge.} – \textit{open-ended datasets.} (a) Knowledge and (b) format following measures via LLM-as-a-judge across 120 tuning steps. All
results are aggregated over 3 open-ended datasets, 7 models and 3 seeds.}
\label{fig:lora_open_format_knowledge_appendix}
\end{figure*}

\begin{figure*}[t]
\centering

\begin{minipage}{0.58\textwidth}
\makebox[\linewidth][l]{\textnormal{(a)}}\\[-2pt]
\includegraphics[width=\linewidth]{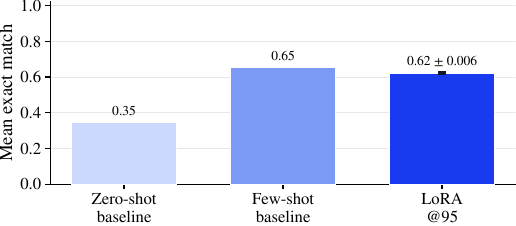}
\end{minipage}
\hfill
\begin{minipage}{0.38\textwidth}
\makebox[\linewidth][l]{\textnormal{(b)}}\\[-2pt]
\includegraphics[width=\linewidth]{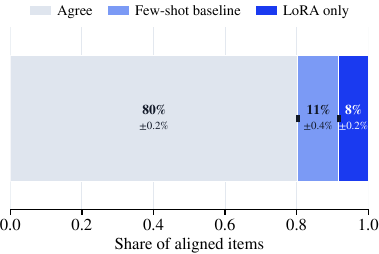}
\end{minipage}
\caption{\textbf{LoRA shows slightly lower aggregated scores compared to few-shot baselines.} On GSM8K, SQuAD, and TriviaQA, LoRA-tuned base models reach slightly lower aggregated scores than the few-shot baseline, with 80\% item-level agreement between the two.}
\label{fig:lora_baseline_alignment}
\end{figure*}

\begin{figure*}[t]
\centering
\begin{minipage}{0.48\textwidth}
\makebox[\linewidth][l]{\textnormal{(a)}}\\[-2pt]
\includegraphics[width=\linewidth]{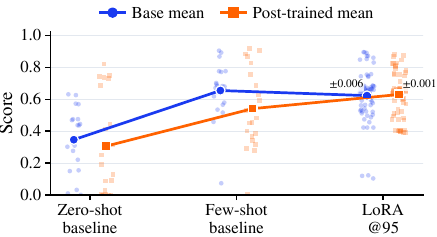}
\end{minipage}
\hfill
\begin{minipage}{0.48\textwidth}
\makebox[\linewidth][l]{\textnormal{(b)}}\\[-2pt]
\includegraphics[width=\linewidth]{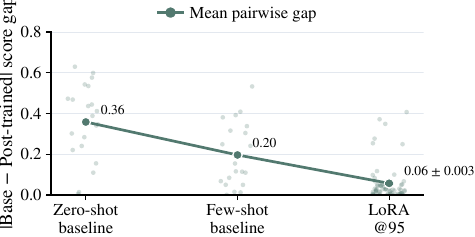}
\end{minipage}

\caption{\textbf{LoRA-tuned base and post-trained models align best in benchmark scores, outperforming zero- and few-shot baselines.} (a) Task accuracy (exact match), aggregated over all datasets and seeds. (b) Average base--post-trained score difference per configuration across all datasets, models, and seeds.}
\label{fig:lora_benchmark_score_appendix}
\end{figure*}

\begin{figure*}[t]
\centering
\begin{minipage}{0.45\textwidth}
\centering
\makebox[\linewidth][l]{\textnormal{(a)}}\\[-2pt]
\includegraphics[width=0.8\linewidth]{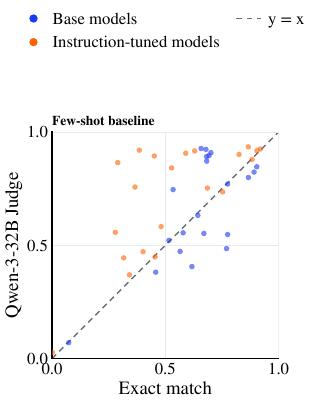}
\end{minipage}
\hfill
\begin{minipage}{0.45\textwidth}
\centering
\makebox[\linewidth][l]{\textnormal{(b)}}\\[-2pt]
\includegraphics[width=0.8\linewidth]{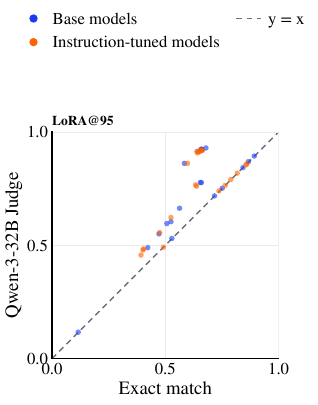}
\end{minipage}
\caption{\textbf{LoRA improves the agreement between Qwen-3-32B judge and exact match scores compared to baselines.} a) Agreement between Qwen-3-32B judge and exact match scores under few-shot baseline answers and b) under LoRA tuned model answers. Similar to the findings in soft-prompt tuning LoRA lets the models output clean answers following the expected format which leads LLM-judge and exact match metrics to align well.}
\label{fig:qwen_exact_match_lora_appendix}
\end{figure*}

We observe that LoRA also maximizes format following without introducing new systematic task knowledge (Fig.~\ref{fig:lora_close_format_knowledge_appendix} and Fig.~\ref{fig:lora_open_format_knowledge_appendix}). On GSM8K, SQuAD, and TriviaQA, LoRA-tuned base models reach slightly lower aggregated scores than the few-shot baseline, with 80\% item-level agreement between the two (Fig.~\ref{fig:lora_baseline_alignment}). Like soft-prompt tuning, LoRA brings base and instruction-tuned benchmark scores closer together than the baselines do (Fig.~\ref{fig:lora_benchmark_score_appendix}), and it improves agreement between the Qwen-3-32B judge and exact-match scoring (Fig.~\ref{fig:qwen_exact_match_lora_appendix}).
Overall, we see strong signal that our findings can extend to other PEFT methods as well.

\subsubsection{Comparison between LoRA and Soft-Prompts}
\label{lora_sp_appendix}

\begin{figure*}[t]
\begin{center}
\includegraphics[width=0.4\linewidth]{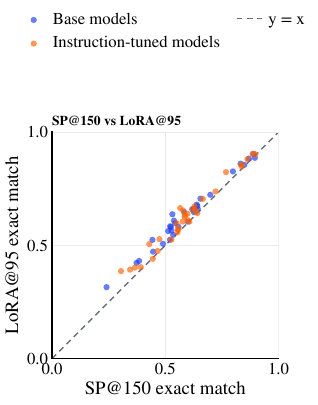}
\end{center}
\caption{\textbf{Soft-prompt and LoRA scores have a linear trend for both base and instruction-tuned models.} -- \textit{Belebele, ChemBench, MMLU, GSM8K, SQuAD, TriviaQA}. Task accuracy alignment between LoRA- and soft-prompt tuned base and post-trained models. Each point shows an average score over 3 seeds. We see that the points create a linear line indicating an alignment between actual task accuracy scores evaluated with LoRA and soft-prompt tuning.}
\label{fig:lora_sp_benchmarks_score_appendix}
\end{figure*}

In this section, we compare the benchmarks scores achieved with soft-prompt tuning and LoRA on both base and post-trained models. Fig.~\ref{fig:lora_sp_benchmarks_score_appendix} shows a scatter plot of all base and post-trained models evaluated under LoRA and soft-prompt tuning. The plot reveals a clear linear trend between soft-prompt and LoRA performance, indicating close alignment between the two methods. %

\clearpage

\subsection{Format-Following and Knowledge Accuracy Examples}
\label{appendix_metric_examples}

\begin{promptbox}[Open-ended -- Knowledge and Format Following Accuracy (GSM8K)]\footnotesize
    \textbf{Question:} Each slice of cheesecake contains 350 calories. If there are
    2800 calories in every cheesecake, and Kiley ate 25\% of the cheesecake, how
    many slices of cheesecake did Kiley eat?

    \textbf{Answer:} 25\% of 2800 calories is 700 calories. 700 calories is 2 slices
    of cheesecake. So Kiley ate 2 slices of cheesecake.

    \textbf{GT:} If there are 2800 calories in every cheesecake, and each slice
    contains 350 calories, then the cheesecake has $2800 / 350 = 8$ slices.
    Kiley ate $8 \times 0.25 = 2$ slices. So the answer is 2.

    \textbf{Knowledge Accuracy:} 1

    \textbf{Format Following Accuracy:} 0
\end{promptbox}

\begin{promptbox}[Closed-Ended - Knowledge and Format Following Accuracy (MMLU)]\footnotesize
    
    \textbf{Question:} A swimmer is able to propel himself forward through the water by moving his arms. Which of the following correctly states the applicant and recipient of the force responsible for the swimmer’s forward acceleration?
    
    \textbf{Possible answers:}
    \begin{itemize}
    \item The force of the surrounding water on the swimmer’s arms
    \item The force of the swimmer’s arms on the swimmer’s torso
    \item The force of the swimmer’s arms on the surrounding water
    \item The force of the swimmer’s torso on the swimmer’s arms
    \end{itemize}
    
    \textbf{Answer:} The force of the swimmer’s arms on the surrounding water
    
    \textbf{GT:} The force of the surrounding water on the swimmer’s arms

    \textbf{Knowledge Accuracy:} 0

    \textbf{Format Following Accuracy:} 1
\end{promptbox}

\newpage

\subsection{Exploratory Multi-Dataset Global Soft-Prompt Analysis}
\label{global_sp_appendix}

We additionally ran a small exploratory experiment to test whether a single soft-prompt per model, trained on a mixture of datasets (\textit{global} soft-prompt), can replace dataset-model specific soft-prompts. We train a 30-token global soft-prompt on a balanced mixture of five open-ended datasets (GSM8K, MATH, DROP, TriviaQA, SQuAD) with batch size 8, and evaluate it on SQuAD, TriviaQA, and DROP for Llama-3.1-8B, Ministral-3-14B, and Qwen-3-8B, in both base and instruction-tuned variants. We compare the global soft-prompt at step 900 with the dataset-specific 10-token soft-prompts at step 150. Under balanced sampling, this corresponds to roughly 180 updates' worth of examples per dataset, so the budgets are approximately matched.

\begin{figure*}[t]
\centering

\includegraphics[width=\linewidth]{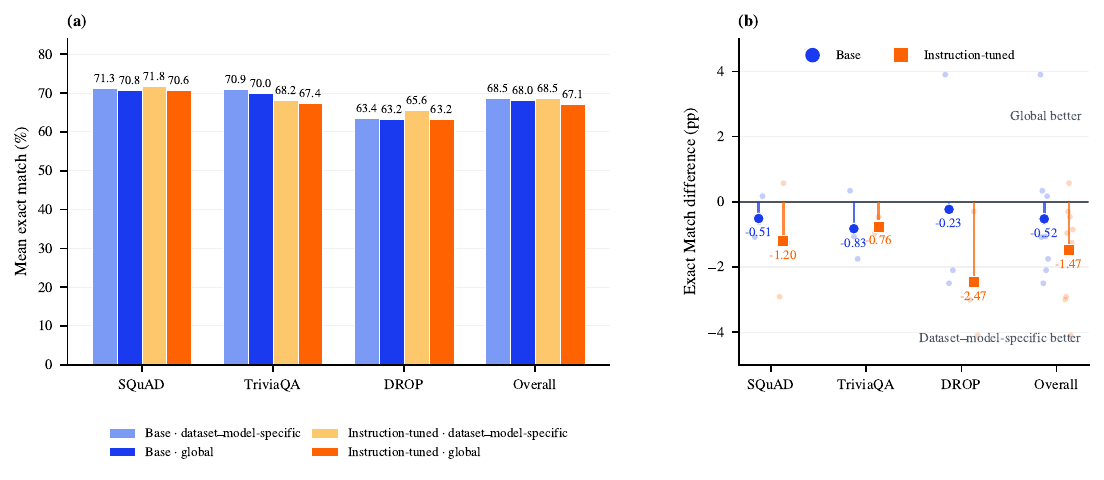}
\caption{\textbf{Global soft-prompt vectors largely preserves the performance of dataset-model specific prompts.}}
\label{fig:global_sp_appendix}
\end{figure*}

The global soft-prompt largely preserves the performance of dataset-model specific prompts. Averaged across models and benchmarks, exact match decreases by 0.52 percentage points (pp) for base models and 1.47 pp for instruction-tuned models, with mean absolute differences of 1.50 pp and 1.60 pp, respectively (Fig.~\ref{fig:global_sp_appendix}).

We believe this experiment shows promising results for further studying the effect of a global soft-prompt. However, broader conclusions would require varying prompt size, dataset composition, model family, and random seed. It is also important to note that dataset-model specific soft-prompt tuning cost is minimal: it requires only 1--2 minutes of training per model and benchmark. Consequently, reduced training cost alone does not establish a practical need for a global soft-prompt.

\subsection{Licenses of the Models, Datasets and Software}
\label{licenses}

Below we list licenses for all the models, datasets and software we have used in this paper. Table~\ref{tab:model-licenses} shows model and software licenses, whereas Table~\ref{tab:dataset-licenses} show the dataset licenses.

\begin{table*}[h]
\centering
\caption{Licenses of models and software used in this paper.}
\label{tab:model-licenses}
\begin{tabular}{ll}
\toprule
\textbf{Model / Software} & \textbf{License} \\
\midrule
Llama-3.1-8B (base \& instruct) \citep{grattafiori_llama_2024}    & Llama 3.1 Community License \\
Llama-3.2-1B (base \& instruct) \citep{grattafiori_llama_2024}    & Llama 3.2 Community License \\
Ministral-3-3B (base \& instruct) \citep{liu_ministral_2026}      & Apache 2.0 \\
Ministral-3-14B (base \& instruct) \citep{liu_ministral_2026}     & Apache 2.0 \\
Olmo-3-7B (base \& instruct) \citep{olmo_olmo_2026}               & Apache 2.0 \\
OLMo-2-32B (LLM-as-a-judge) \citep{olmo_2_2025}                   & Apache 2.0 \\
Qwen3-4B (base \& instruct) \citep{yang_qwen3_2025}               & Apache 2.0 \\
Qwen3-8B (base \& instruct) \citep{yang_qwen3_2025}               & Apache 2.0 \\
Qwen3-32B (LLM-as-a-judge) \citep{yang_qwen3_2025}                & Apache 2.0 \\
\midrule
OLMES \citep{gu_olmes_2025}                                       & Apache 2.0 \\
\bottomrule
\end{tabular}
\end{table*}

\begin{table*}[h]
\centering
\caption{Licenses of datasets used in this paper.}
\label{tab:dataset-licenses}
\begin{tabular}{lll}
\toprule
\textbf{Dataset} & \textbf{Type} & \textbf{License} \\
\midrule
TriviaQA \citep{joshi_triviaqa_2017} & Open-ended    & Apache 2.0 \\
GSM8K \citep{cobbe_training_2021}              & Open-ended    & MIT \\
SQuAD \citep{rajpurkar_squad_2016}   & Open-ended    & CC BY-SA 4.0 \\
DROP \citep{dua_drop_2019}                      & Open-ended  & CC BY-SA 4.0 \\
MATH \citep{hendrycks_measuring_2021}           & Open-ended  & MIT \\
QASPER \citep{dasigi_dataset_2021}              & Generative   & CC BY 4.0 \\
QuALITY \citep{pang_quality_2022}               & Generative   & CC BY 4.0 \\
ContractNLI \citep{koreeda_contractnli_2021}    & Generative   & CC BY 4.0 \\
MMLU \citep{hendrycks_measuring_2020}       & Closed-ended  & MIT \\
Belebele \citep{bandarkar_belebele_2024} & Closed-ended & CC BY-SA 4.0 \\
ChemBench \citep{mirza_framework_2025}      & Closed-ended  & MIT \\
\bottomrule
\end{tabular}
\end{table*}

\end{document}